\documentclass[3p,12pt, authoryear,english]{elsarticle}

\usepackage[english]{babel}
\usepackage[utf8]{inputenc}
\usepackage{graphicx}
\usepackage{url}
\usepackage{datetime}
\usepackage{lineno}
\usepackage{amsmath}
\usepackage{todonotes}
\usepackage{booktabs}
\usepackage{algorithm} 
\usepackage{algpseudocode}
\usepackage{breakcites}
\usepackage{pifont}
\usepackage{xcolor}
\usepackage[labelsep=period]{caption}
\usepackage[caption=false]{subfig}

\input amssym.tex
\begin{document}

\begin{frontmatter}
\title{Improving the Accuracy of Global Forecasting Models using Time Series Data Augmentation}
%\author{}
\author[monash]{Kasun~Bandara \corref{cor1}}%\corref{cor1}
\author[monash]{Hansika~Hewamalage}
\author[monash]{Yuan-Hao~Liu}
\author[beihang]{Yanfei~Kang}
\author[monash]{Christoph~Bergmeir}

\address{Herath.Bandara@monash.edu, Christoph.Bergmeir@monash.edu, Hansika.Hewamalage@monash.edu, yanfeikang@buaa.edu.cn}
\address[monash]{Faculty of Information Technology, Monash University, Melbourne, Australia.}
\address[beihang]{School of Economics and Management, Beihang University, Beijing  100191, China.}

\cortext[cor1]{Postal Address: Faculty of Information Technology, P.O. Box 63 Monash University, Victoria 3800, Australia. E-mail address: Herath.Bandara@monash.edu}

\begin{abstract}
Forecasting models that are trained across sets of many time series, known as Global Forecasting Models (GFM), have shown recently promising results in forecasting competitions and real-world applications, outperforming many state-of-the-art univariate forecasting techniques. In most cases, GFMs are implemented using deep neural networks, and in particular Recurrent Neural Networks (RNN), which require a sufficient amount of time series to estimate their numerous model parameters. However, many time series databases have only a limited number of time series. In this study, we propose a novel, data augmentation based forecasting framework that is capable of improving the baseline accuracy of the GFM models in less data-abundant settings. We use three time series augmentation techniques: GRATIS, moving block bootstrap (MBB), and dynamic time warping barycentric averaging (DBA) to synthetically generate a collection of time series. The knowledge acquired from these augmented time series is then transferred to the original dataset using two different approaches: the pooled approach and the transfer learning approach. When building GFMs, in the pooled approach, we train a model on the augmented time series alongside the original time series dataset, whereas in the transfer learning approach, we adapt a pre-trained model to the new dataset. In our evaluation on competition and real-world time series datasets, our proposed variants can significantly improve the baseline accuracy of GFM models and outperform state-of-the-art univariate forecasting methods.
\end{abstract}

\end{frontmatter}

\section{Introduction}
\label{sec:intro}

In many industries, such as retail, food, railway, mining, tourism, energy, and cloud-computing, generating accurate forecasts is vital as it provides better grounds for decision-making of organisational short-term, medium-term and long-term goals. Here, industrial application databases often consist of collections of related time series that share key features in common. To generate better forecasts under these circumstances, recently, global forecasting models (GFM) have been introduced as a competitive alternative to the traditional univariate statistical forecasting methods \citep{Januschowski2020-ud}, such as exponential smoothing \citep{Hyndman2008-yd} and autoregressive integrated moving average \citep[ARIMA,][]{Box2015-bz}. Compared to univariate forecasting methods that treat each time series separately and forecast each series in isolation, GFMs are unified forecasting models that are built using all the available time series. And thus the GFMs are able to simultaneously learn the common patterns available across a rich collection of time series, and offer much better scalability to the increasing volumes of time series. In particular, deep learning based GFMs have recently achieved promising results by outperforming many state-of-the-art univariate forecasting methods \citep{Smyl2019-cb, Lai2018-zx, Borovykh2017-vz, Salinas2019-dl, Wen2017-ky, Bandara2019-iv, Hewamalage2019-il, Bandara2020-zt,kang2020tcn,Bandara2020-en}.
 
The GFMs outshine univariate forecasting models, in situations where large quantities of related time series are available from the same domain \citep{Januschowski2020-ud}. Examples include the sales demand of related product assortments in retail, the ride-share services demand in multiple regions, server performance measures in computer centres, household smart meter data, and others. The requirement of having adequate amounts of related time series becomes essential when building accurate GFMs, as the model parameters are estimated jointly using all the available time series. This requirement becomes vital for deep learning based GFMs, as they are inherently data ravenous, and require large numbers of model parameters to be estimated. However, in situations where time series databases are constrained by the amounts of time series available, i.e., small to medium sized datasets, GFMs may not reach their full potential in accuracy.

In the absence of adequate amounts of time series data, GFMs may not be able to learn important characteristics of time series, such as seasonality. In such circumstances, one approach is to supplement the model training procedure by incorporating expert knowledge available about time series. In a situation where such expert knowledge is not readily and explicitly available, data augmentation (DA) techniques can be used to artificially generate new copies of data to increase the sample size in use, and thereby enable the model to learn various aspects of the data better. 
Here, the DA approach addresses the data sparsity issue by generating synthetic data to increase the number of observations available for model training. The application of the DA strategy has proven successful in various machine learning applications, such as image classification \citep{Krizhevsky2012-ct}, speech recognition \citep{Hannun2014-ju, Wang2020-pc}, text classification \citep{Zhang2015-vg}, general semi-supervised classification \citep{DONYAVI2020107543}, and time series related research \citep{Forestier2017-su, Bergmeir2016-zk, Fawaz2018-fj,Kang2019-dy,talagala2019bmsr}. Another approach to overcome the extensive data requirements of learning algorithms is by transferring the knowledge representations from a background dataset to a target dataset. This process is commonly referred to as transfer learning (TL) in the machine learning literature. In TL, a base model is initially trained using a background or a source dataset that models the source task. The pre-trained model is then transferred to a target dataset with a target task. The TL strategy is particularly useful when the target dataset is significantly smaller than the background dataset. As a result, TL based approaches have been used in a wide range of machine learning applications, such as image classification \citep{Yu2017-dm,Yosinski2014-xz,Huang2020-fv,Zhuang2018-nb,Li2020-og}, language modelling \citep{Purushotham2017-sz,Yoon2017-tf,Glorot2011-kp}, and more recently in imaging based time series forecasting \citep{li2020imaging}. The success of TL based approaches in these applications can be mainly attributed to the rich data structure inherently available in text, images, and speech related data. As a result, pre-trained models are able to capture the common features of data that can be easily transferable to the target task, while obviating the target task to learn the general characteristics of data from scratch.

When using DA in the time series context, one branch of techniques aims to generate artificial time series that are similar to the data generation process (DGP) of the original dataset \citep{Forestier2017-su, Bergmeir2016-zk}, whereas other techniques generate random sets of time series that may not be similar to the DGP of the original dataset, and they only resemble general characteristics of time series. When building GFMs, the knowledge of the augmented time series can be transferred to a target dataset in two different ways: training a GFM by pooling the augmented time series and the original time series together or pre-training a GFM using the augmented time series, and therewith transferring the knowledge representations of the pre-trained model to the original dataset using the TL strategy. While several studies have investigated the use of these approaches for time series forecasting, a thorough study has not yet been explored in the GFM context. As the recent success of GFMs mostly depends on data-abundant settings \citep{Smyl2019-cb,Salinas2019-dl, Bandara2019-bg}, it is crucial to investigate the use of GFMs with limited time series data, and to develop strategies to make GFMs competitive under these circumstances. Motivated by this gap, in this study, we propose a GFM based forecasting framework that can be used to improve the forecast accuracy in data-sparse time series databases. As the primary prediction module of our framework, we use Recurrent Neural Networks (RNN), a promising neural network (NN) architecture that has been heavily used in the recent GFM literature \citep{Smyl2019-cb, Salinas2019-dl, Bandara2019-iv, Hewamalage2019-il, Bandara2020-zt}.

In this study, we demonstrate the use of DA techniques to improve the accuracy of GFMs in data-sparse environments. We use three different DA techniques to synthetically generate time series, namely: 1)  GeneRAting TIme Series with diverse and controllable characteristics \citep[GRATIS,][]{Kang2019-dy}, 2) moving block bootstrap \citep[MBB,][]{Bergmeir2016-zk}, and 3) dynamic time warping barycentric averaging \citep[DBA,][]{Forestier2017-su}. GRATIS \citep{Kang2019-dy} is a statistical generative model that artificially generates time series with diverse characteristics, which are not necessarily similar to the DGP of the original dataset. Whereas, the MBB and the DBA methods are aimed to generate time series that are similar to the DGP of the original dataset. As described earlier, we transfer the knowledge representation of the augmented time series to the original dataset using two different approaches. In the first approach, we pool the synthetically generated time series together with the original dataset, and build a GFM across all the available time series (pooled strategy). In our second approach, we pre-train a GFM using the augmented time series, and thereafter transfer the knowledge representations of the pre-trained models to the original dataset using the TL methodology (transfer strategy). Here, we use different TL schemes to import the information from the augmented data to a target dataset. Based on the above strategies, we propose a set of model variants to evaluate against the baseline model, which is built using the original set of time series. Furthermore, we use state-of-the-art statistical forecasting techniques as benchmarks to compare against our proposed methods. The proposed forecasting framework is attested using five time series databases, including two competition datasets and three real-world datasets.

The rest of the paper is organised as follows. In Section~\ref{sec:relatedwork}, we provide a brief review on TL techniques and their recent applications to time series forecasting, and an overview of time series DA approaches. In Section~\ref{sec:framework}, we discuss the architecture of our forecasting engine in detail. The proposed TL schemes are described in Section~\ref{sec:transferlearning}. In Section~\ref{sec:dataaugmentation}, we explain the time series augmentation techniques used in this study. Our experimental setup is discussed in Section~\ref{sec:experiments}. Finally, we conclude our paper in Section~\ref{sec:con}.

\section{Related Work}
\label{sec:relatedwork}

In the following, we discuss the related work in the areas of DA and TL for time series analysis.

\subsection{Time series augmentation}
The application of DA techniques in machine learning models has seen success in many domains. \cite{Chen2018-wo} use a Bayesian generative adversarial network to generate new samples of wind and solar energy input data with different variations. Whereas, \cite{Esteban2017-yg} use recurrent generative adversarial networks to generate synthetic clinical records in the medical domain, where ethical restrictions often constrain the data collection. Similar to TL, the successful application of DA can be seen in image classification \citep{Krizhevsky2012-ct}, speech recognition \citep{Hannun2014-ju}, and text classification \citep{Zhang2015-vg}.

There exists a large body of literature that discusses various forms of DA techniques available for time series analysis. This includes time series bootstrapping methods \citep{Bergmeir2016-zk, Iftikhar2017-wy}, time series averaging techniques \citep{Forestier2017-su}, and statistical generative models \citep{Denaxas2015-sl, Papaefthymiou2008-va, Kegel2018-rs, Kang2019-dy}. \cite{Bergmeir2016-zk} use the MBB method to generate multiple versions of a given time series, and build an ensemble of exponential smoothing forecasting models on the augmented time series. \cite{Forestier2017-su} use DBA, a data augmentation strategy that averages a set of time series to produce new samples of data for time series classification. Markov Chain Monte Carlo (MCMC) techniques are also frequently used in the literature to generate synthetic time series \citep{Denaxas2015-sl, Papaefthymiou2008-va}. More recently, \cite{Kang2019-dy} propose GRATIS that uses mixture autoregressive models to generate time series with diverse and controllable characteristics.

On the other hand, recent studies also investigate the use of NNs as a time series augmentation technique \citep{Almonacid2013-ot,Le_Guennec2016-tc}. \cite{Almonacid2013-ot} generate a set of ambient temperature hourly time series using a Multi-Layer Perceptron architecture, and demonstrate the effectiveness of using NNs to generate time series closer to the real-world data. Furthermore, \cite{Le_Guennec2016-tc} employ a deep Convolutional NN architecture to generate synthetic data for time series classification. More recently, deep NN based Generative Adversarial Network (GAN) architectures \citep{Goodfellow2014-kb} have received significant attention in the area of DA. The competition-driven training mechanism employed in GANs, allows the network to generate realistic samples, similar to the DGP of the source dataset. In the literature, though GAN architectures are used mostly for image generation, more recent studies have shown that GANs can also be applied to generate new copies of time series \citep{Fu2019-mw,Esteban2017-yg,Zhang2018-xn,Yoon2019-gr}.

\subsection{Transfer learning}
\label{sec:transferlearningwork}

The existing TL methods can be distinguished by the type of knowledge representation to be transferred and how these representations are transferred. In the following, $T_s$ denotes the source or background task, $D_s$ the corresponding dataset, $T_t$ the target task, and $D_t$ the corresponding dataset. The TL approaches can be mainly categorised into three types, based on the different transfer methods between $T_s$ and $T_t$ \citep{Pan2010-fi}, namely: Inductive TL, Transductive TL and Unsupervised TL. The Inductive TL is typically used when $T_t$ is different from $T_s$, irrelevant of their domains.
The Transductive TL is applied when $T_t$ and $T_s$ are the same, while their respective domains are different. According to \cite{Pan2010-fi}, Transductive TL can be further categorised based on the similarities between the feature spaces of $D_s$ and $D_t$. Finally, the unsupervised transfer learning is used when the labelled data are not available in $D_s$ and $D_t$ for model training. Furthermore, these approaches can be used to transfer various forms of knowledge representations available in $D_s$.
For example, the instance-transfer approach aims to reuse certain parts of $D_s$ for the learning tasks of $T_t$ by applying instance re-weighting. The feature-transfer approach attempts to transfer the knowledge of $D_s$ in the form of feature representation, whereas in the parameter-transfer approach, knowledge is represented by the model parameters or prior distributions that may be shared between the $T_s$ and $T_t$. The relational knowledge-based transfer is expected to exploit similar relationships among $D_s$ and $D_t$. For further discussions and definitions of these TL paradigms, we refer to \cite{Pan2010-fi}.

To overcome the data ravenousness of modern-day deep learning algorithms, the TF based approaches have been introduced in many domains, such as image classification \citep{Yu2017-dm,Yosinski2014-xz,Bengio2012-tr} and language modelling \citep{Purushotham2017-sz,Yoon2017-tf,Glorot2011-kp,Ramachandran2017-px}. With respect to image classification, \cite{Bengio2012-tr} investigate the preliminary results of using transfer learning on images, and \cite{Yosinski2014-xz} explore the use of shared parameters between the source and target domains to improve the accuracy in image classification tasks. Those authors also argue that the initial layers in an NN model tend to capture the general features of an image, while the last layers aim to embed more specific features. In terms of language modelling, \cite{Glorot2011-kp} propose a feature-transfer approach that uses a stacked denoising autoencoder to learn the invariant representation between the source and target domains. It allows the sentiment classifiers to be trained and deployed on different domains. Furthermore, \cite{Yoon2017-tf} compare various TL schemes available for personalised language modelling using RNNs. In the TL process, those authors control the number of trainable parameters of the target model by freezing the initial layers of the RNNs. Also, \cite{Purushotham2017-sz} use variational RNNs to capture underlying temporal latent dependencies in language models, whereas \cite{Ramachandran2017-px} implement a parameter-transfer approach to use pre-trained weights of the base model to initialise the target model.

More recently, the application of TL methods is also gaining popularity in time series forecasting research. \cite{Ribeiro2018-aa} introduce Hephaestus, a TL based forecasting framework for cross-building energy prediction, to improve the accuracy of energy estimations for new buildings with limited historical data. There, those authors propose a seasonal and trend adjusted approach that allows Hephaestus to transfer knowledge across similar buildings with different seasonal and trend profiles. The research work in \cite{Laptev2018-kb} proposes a loss function to reconstruct the input data of the model, and thereby extract time series features using a stack of fully connected LSTM layers. Those authors show that this feature-transfer approach leads to significant accuracy improvements over the traditional TL approaches, in situations where the size of the target dataset is small. To handle time-varying properties in time series data, \cite{Ye2018-ex} propose a hybrid algorithm, based on TL that effectively accounts for the observations in the distant past, and leverages the latent knowledge embedded in past data to improve the forecast accuracy. Moreover, \cite{Gupta2018-no} implement an RNN autoencoder architecture to extract generic sets of features from multiple clinical time series databases. Those features are then used to build simple linear models on limited labelled data for multivariate clinical time series analysis. \citet{li2020imaging} first transform time series into images and use TL for image feature extraction. The extracted features are used as time series features to obtain the optimal weights of forecast combination \citep{Kang2019-dy}.

In summary, we have identified feature-transfer learning and parameter-transfer learning approaches as the most commonly used TL paradigms in deep learning based applications. It can be mainly attributed to the capability of NNs to extract non-trivial latent representations of data.

\section{Forecasting Framework}
\label{sec:framework}
In this section, we describe in detail the main components of our proposed GFM based forecasting framework. The framework consists of three layers, namely: 1) the pre-processing layer, 2) the RNN training layer, and 3) the post-processing layer. In the following, we first discuss the pre-processing techniques used in our forecasting framework. Then, we provide a brief introduction to residual RNNs, which are the primary prediction unit of our forecasting engine. Finally, we explain the functionality of the post-processing layer of the framework.

\subsection{Time series pre-processing}
\label{sec:pre-processing}

As GFM methods are trained across a group of time series, accounting for various scales and variances present in these time series becomes necessary \citep{Hewamalage2019-il,Bandara2019-iv}. Therefore, as the first step in our pre-processing pipeline, we normalise the collection of time series $\mathcal X = \{X_i\}_{i = 1}^N$ using a \emph{meanscale} transformation strategy \citep{Salinas2019-dl, Hewamalage2019-il}, which can be defined as follows:

\begin{equation}
X_{i, \mathrm{normalised}} = \frac{X_i}{\frac{1}{k}\sum_{t=1}^{k}{X_{i,t}}},
\label{meanscale}
\end{equation}
where $X_{i, \mathrm{normalised}}$ represents the $i$th normalised time series, and $k$ is the number of observations in time series $X_i$, where $i \in \{1, 2, \cdots, N\}$.

We then stabilise the time series' variance by log transformation. It also allows us to convert possible multiplicative seasonal and trend components of a given time series into additive ones, which is necessary for the last step in our pre-processing pipeline, the deseasonalisation process. To avoid problems for zero values, we use the log transformation in the following way:

\begin{equation}
X_{i, \mathrm{normalised}~\&~\mathrm{logscaled}} = \begin{cases}
  \log(X_{i, \mathrm{normalised}}), & min(\mathcal X)>0;\\
  \log(X_{i, \mathrm{normalised}} + 1), & min(\mathcal X)=0,\\
\end{cases}
\label{logscale}
\end{equation}
where $\mathcal X$ denotes the full set of time series, and $X_{i, \mathrm{normalised}~\&~\mathrm{logscaled}}$ is the corresponding normalised and log transformed time series of $X_i$. Note that we assume the time series to forecast are non-negative. 

As the next step of our pre-processing pipeline, we introduce a time series deseasonalisation phase to extract the seasonal components from time series. Following \cite{Bandara2020-zt}, when using NNs for forecasting, these extracted seasonal components can be used in two different ways. In the first training paradigm, those authors suggest the Deseasonalised (DS) approach, which removes the extracted seasonal values from a time series, and then uses the remainder, i.e., trend and residual components, to train the NN. Here, as the seasonal components are removed from the time series, an additional reseasonalisation step is introduced in the post-processing phase to predict the future seasonal values of the time series. In the second training paradigm, Seasonal Exogenous (SE) approach, the extracted seasonal components are used as exogenous inputs in addition to the original observations of the time series. As the time series are not seasonally adjusted in this approach, an additional reseasonalisation step is not required in the post-processing phase. The main objective of these two training paradigms is to supplement the subsequent NN’s learning process. Those authors suggest that the accuracy of these two variants depends on the seasonal characteristics of the time series. In line with the recommendations by \cite{Bandara2020-zt}, we use these two approaches accordingly in our experiments. In Section~\ref{sec:experiments}, we summarise the training paradigms used for each dataset.

Following the recent success of Seasonal-Trend Decomposition~\citep[STL,][]{Cleveland1990-rc} as a pre-processing technique for NNs \citep{Bandara2019-iv, Hewamalage2019-il, Bandara2020-zt}, we use it to extract the seasonal components from time series. When we apply STL to a normalised and log scaled time series $X_{i,\mathrm{normalised}~\&~\mathrm{logscaled}}$, its additive decomposition can be formulated as follows: 
\begin{equation}
X_{i,\mathrm{normalised}~\&~\mathrm{logscaled}} = \hat{S}_i + \hat{T}_i  + \hat{R}_i,
\label{additive}
\end{equation}
where $\hat{S}_i $, $\hat{T}_i$, $\hat{R}_i$ are the corresponding seasonal, trend, and the residual components of the time series $X_{i,\mathrm{normalised}~\&~\mathrm{logscaled}}$, respectively. In this study, we use the R \citep{R_Core_Team2013-bo} implementation of the STL algorithm, \verb|stl|, from the \verb|forecast| package \citep{Hyndman2015-vm,Khandakar2008-hd}.

%Following the recommendations of \cite{Bandara2019-iv,Hewamalage2019-ko}, we configure the \verb|s.window| parameter to ``periodic'' in STL to ensure extracting only the deterministic form of seasonality ($\hat{S}^{d}_i$) from a time series. The remaining stochastic seasonality ($\hat{S}^{s}_i$) is included as seasonal lags to train our network (see Section~\ref{sec:rnn}). \cite{Bandara2019-iv} argue that this approach can be seen as a ‘boosting’ ensemble technique \citep{Schapire2003-xb}, where the deseasonalisation process is analogous to a weak base learner that supplements the RNN learning procedure in our forecasting engine.

\subsection{Residual Recurrent Neural Networks}
\label{sec:rnn}

Nowadays, the application of deep learning models is gaining popularity among the time series forecasting community \citep{Smyl2019-cb, Stepnicka2016-bu,Borovykh2017-vz, Bandara2020-zt}. Many of these innovations are based on RNN architectures, which were motivated mainly by the continued success of RNNs in modelling sequence related tasks \citep{Mikolov2010-rb,Sutskever2014-vp}. A host of different RNN architectures for time series forecasting exists in the forecasting literature, overviews and discussions \citep{Hewamalage2019-il, Zimmermann2012-cp}.
Based on the recommendations given by \cite{Hewamalage2019-il}, when forecasting with GFMs, we select the Long Short-Term Memory network (LSTM) as our primary RNN architecture and implement the \textit{Stacking Layers} design pattern to train the network. Furthermore, we introduce residual connections to the stacking architecture to address the vanishing gradient problem that may occur in situations with a higher number of hidden layers \citep{Bengio1994-oq}. This was originally proposed by \cite{He2016-wm} as the \textit{Residual Net (ResNet)}, where the authors use residual connections to accommodate substantially deeper architectures of Convolutional NNs (CNN) for image classification tasks. They also argue that learning the residual mappings is computationally easier than directly learning to fit the underlying mapping between input and output. More recently, a variant of the ResNet architecture has been applied to time series forecasting using RNNs \citep{Smyl2019-cb, Smyl2016-ux}. In fact, the residual architecture proposed by \cite{Smyl2019-cb} became the winning solution of the M4 forecasting competition \citep{Makridakis2018-cm}. During the transfer learning phase (see Section~\ref{sec:transferlearning}), we expect to add extra stacking layers to the base architecture. Therefore, having residual connections among the stacking layers becomes necessary for a stable learning process in our network.

\begin{figure}[htbp]
\centerline{\includegraphics[width=0.44\textwidth]{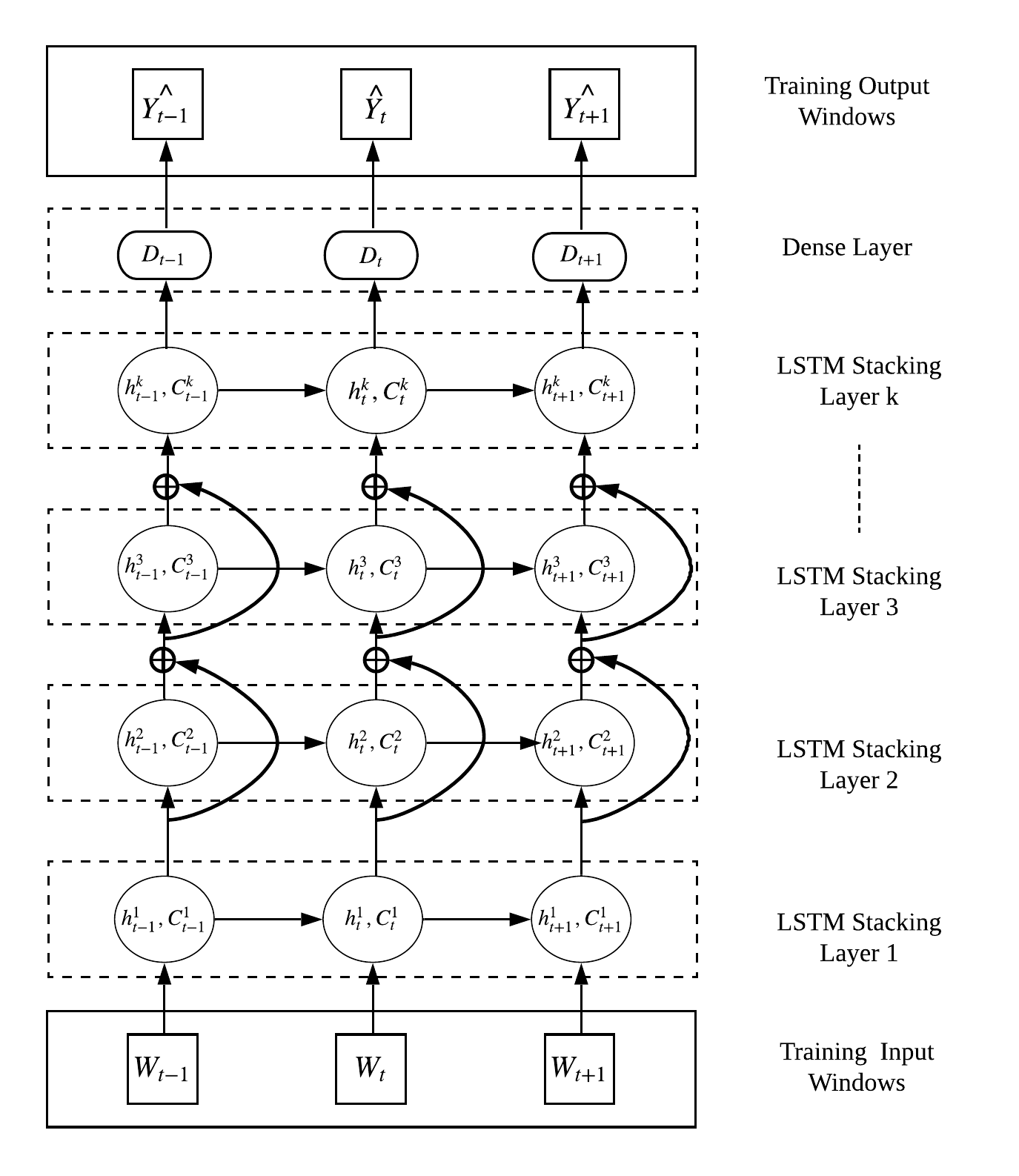}}
\caption{The unrolled representation of a residual recurrent network architecture with an amount of $k$ stacking layers. Here, the residual connections are represented by curved arrows. According to \cite{He2016-wm}, these residual connections allow the stacking layers to fit a residual mapping between $W_t$ and $\hat{Y_{t}}$, while avoiding the network to degrade with increasing network depth.}
\label{forecastingarch}
\end{figure}

Figure~\ref{forecastingarch} illustrates the architecture of the forecasting engine used in our experiments, which mainly consists of three components: an input layer, stacking layers with a dense layer, and an output layer. To train this network, we use the pre-processed time series in the form of input and output training windows. Here, the values of the pre-processed time series depend on the training paradigm, i.e., the DS or SE strategy, used in a particular dataset.
These training windows are generated by applying the Moving Window (MW) transformation strategy to pre-processed time series. Existing studies often recommend the MW strategy, when training NNs for time series forecasting \citep{Smyl2019-cb,Bandara2019-iv,Hewamalage2019-il}. This is mainly due to the Multi-Input Multi-Output (MIMO) principle used in this strategy, where the size of the training output window, $m$ is identical to the size of the intended forecasting horizon $M$. 
In this way, the network is trained to directly predict the entire forecasting horizon $X^M_{i}$ at once, avoiding prediction error accumulation at each forecasting step \citep{Ben_Taieb2012-re}.  Furthermore, on these training windows, we use the local normalisation strategy suggested by \cite{Bandara2019-iv,Hewamalage2019-il} to avoid possible network saturation effects that occur in NNs. Here, the local normalisation strategy used for the DS approach differs from the SE approach. In the DS approach, we use the trend component of the input window's last value, while the mean value of each input window is used in the SE approach.
In Figure~\ref{forecastingarch}, $W_{t}\in{\Bbb R^{n}}$ represents the teacher input window at time step $t$, whereas $\hat{Y_{t}}\in{\Bbb R^{m}}$ represents the LSTM output at time step $t$. Here, $n$ denotes the size of the input window. Moreover, $h_t$ refers to the hidden state of LSTM at time step $t$, while its cell memory at time step $t$ is given by $C_t$.
A fully connected layer $D_t$ (excluding the bias component) is introduced to map each LSTM cell output $h_t$ to the dimension of the output window $m$, equivalent to $ M $. Given the length of the time series $X_i$ as $p$, we use an amount of $(p - m)$ data points from the pre-processed $X_i$ to train our network and reserve the last output window of the pre-processed $X_i$ for the network validation. The L1-norm is used as the primary learning objective function of our training architecture, along with an L2-regularisation term to minimise possible overfitting in the network.

Even though we use the LSTM as the primary RNN cell in this study, we note that our forecasting engine can be used with any other RNN variant such as Elman RNN \citep{Elman1990-my}, Gated Recurrent Units~\citep[GRU,][]{Cho2014-vr}, and others.

\subsection{Post-processing}
We compute the final predictions of our forecasting framework by applying a reseasonalisation process and a denormalisation process to the output given by the LSTM. As outlined earlier in Section~\ref{sec:pre-processing}, the reseasonalisation step is only required when the network is trained using the DS strategy. The reseasonalisation includes forecasting the seasonal components, which have been removed during the pre-processing phase. This is straightforwardly done by copying the last seasonal components of the time series into the future, up to the intended forecast horizon. With respect to denormalisation, we first add the local normalisation factor used in each training window and then reverse the log transform using an exponential function. 
Finally, we multiply the back-transformed vector by the mean of the time series, which is the scaling factor used for the normalisation process.

\begin{figure*}[ht]
\centering
\subfloat[TL.Dense]{
\includegraphics[width=0.26\textwidth]{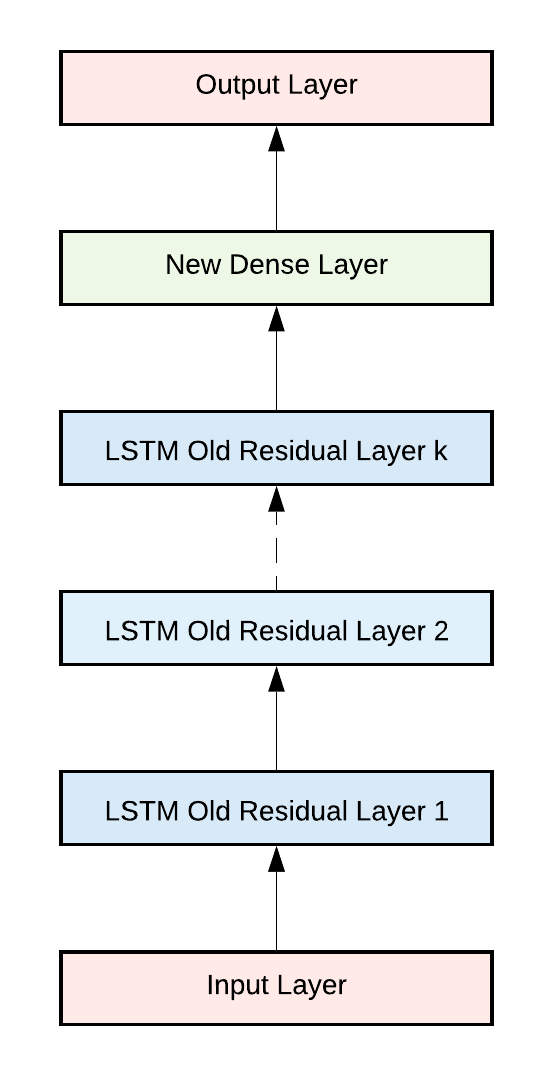}
\label{fig:subfig1}}%\hfill
\subfloat[TL.AddDense]{
\includegraphics[width=0.26\textwidth]{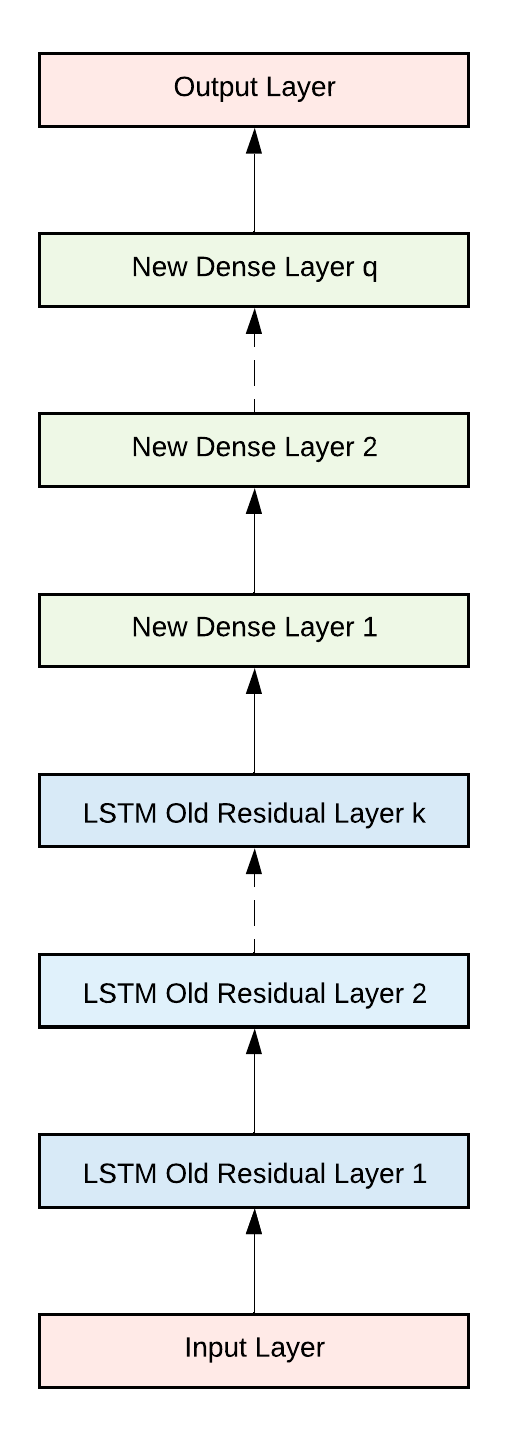}
\label{fig:subfig2}}%\\[-2ex]
\subfloat[TL.LSTM]{
\includegraphics[width=0.26\textwidth]{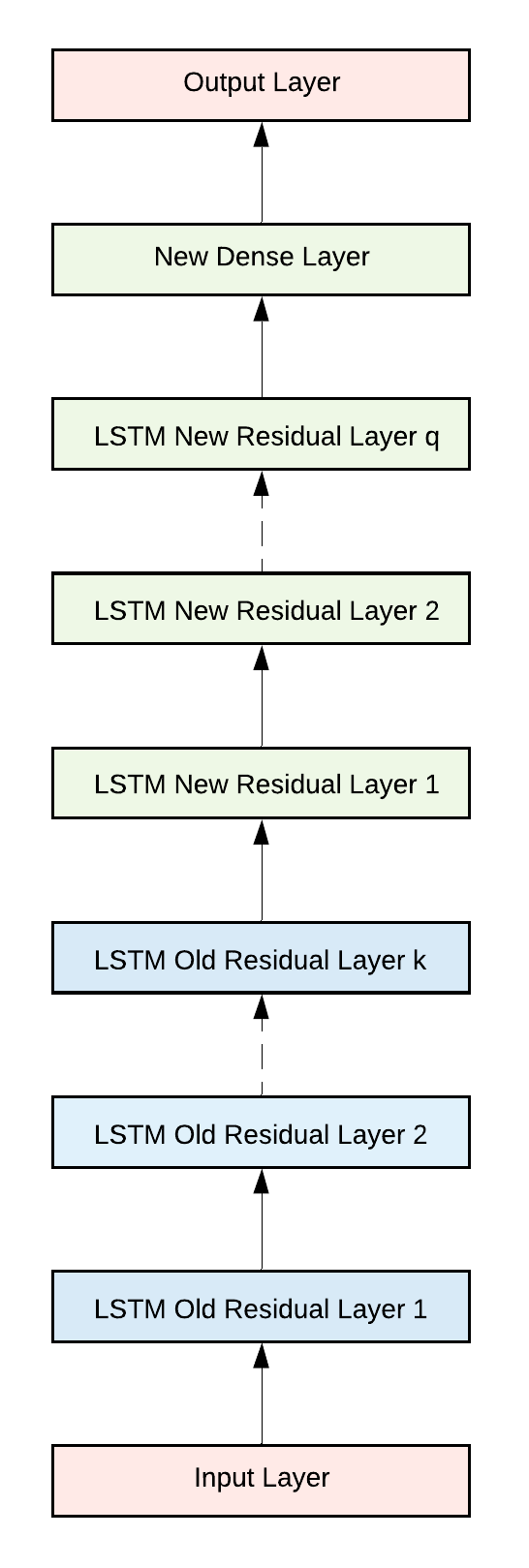}
\label{fig:subfig3}}%\hfill
\caption{An overview of the proposed TL schemes used in this study. The layers used to build the base model using the $D_s$, are represented in blue colour, while the additional layers introduced when building the target model using the $D_t$, are represented in green colour.}
\label{fig:transfarch}
\end{figure*}

\section{Transfer Learning Architectures}
\label{sec:transferlearning}

As discussed in Section~\ref{sec:relatedwork}, \cite{Yoon2017-tf} suggest a host of different TL schemes, when using RNNs for personalised language modelling. In line with these recommendations, we investigate the use of three transfer learning schemes for time series forecasting with the LSTM stacking architecture. Figure~\ref{fig:transfarch} shows the different TL schemes used in this study. Here, we use an abstract view of the proposed three layered residual recurrent architecture (see Figure~\ref{forecastingarch}) to simplify the illustration of the proposed TL schemes. In summary, TL.Dense introduces a dense layer to the pre-trained base model, mapping the output of the base model to the dimension of the output window in $D_t$. TL.AddDense adds an amount of $q$ dense layers to the pre-trained base model. Finally, in TL.LSTM, an amount of $q$ LSTM residual layers with a dense layer is introduced to the base model. In these TL schemes, we assume there exist $k$ residual layers in the base model. We further introduce variants to these TL schemes by changing the total number of trainable parameters in the architecture. Here, we achieve this by training the network, only using the parameters of newly introduced hidden layers, while freezing the hidden layers of the pre-trained base model. Based on these TL schemes, we define the proposed TL architectures as follows:

\begin{description}
\item[TL.Dense.Freeze:] The TL architecture that uses the TL.Dense scheme, while freezing the initial layers of the pre-trained model, and training only the newly added layers.
\item[TL.Dense.Retrain:] The TL architecture that uses the TL.Dense scheme, while re-training initial layers of the pre-trained model and newly added layers.
\item[TL.AddDense.Freeze:] The TL architecture that uses the TL.AddDense scheme, while freezing the initial layers of the pre-trained model, and training only the newly added layers.
\item[TL.AddDense.Retrain:] The TL architecture that uses the TL.AddDense scheme, while re-training initial layers of the pre-trained model and newly added layers.
\item[TL.LSTM.Freeze:] The TL architecture that uses the TL.LSTM scheme, while freezing the initial layers of the pre-trained model, and training only the newly added layers.
\item[TL.LSTM.Retrain:] The TL architecture that uses the TL.LSTM scheme, while re-training initial layers of the pre-trained model and newly added layers.
\end{description}

Here, TL.Dense.Retrain, TL.AddDense.Retrain, and TL.LSTM.Retrain re-train all the layers of the model, whereas the TL.Dense.Freeze, TL.AddDense.Freeze, and TL.LSTM.Freeze re-train only newly added layers to the model. Also, according to the definitions of \cite{Pan2010-fi}, our approach follows the Transductive TL approach (see Section~\ref{sec:transferlearningwork}), where $T_t$ and $T_s$ are the same, i.e., time series forecasting, but the datasets used to train the tasks are different. We implement the above TL learning schemes and the residual RNN architecture proposed in Section~\ref{sec:rnn} using TensorFlow, an open-source deep learning toolkit \citep{Abadi2016-rr}.

\section{Time Series Augmentation}
\label{sec:dataaugmentation}

As highlighted in Section~\ref{sec:intro}, DA techniques are useful for synthetically increasing the number of training samples in a dataset. In this study, we use several time series based DA techniques to artificially generate time series, namely GRATIS, MBB, and DBA. The MBB \citep{Bergmeir2016-zk} and DBA~\citep{Forestier2017-su} methods are expected to generate time series from a similar DGP with that of the original series, whereas the GRATIS method generates time series with diverse characteristics from different DGPs, not related to the original dataset. In the following, we briefly describe the above methods, and explain how we exactly use them in our experiments.

\subsection{GRATIS}

We use GRATIS, a statistical generative model proposed by \cite{Kang2019-dy} to create new time series with diverse characteristics. GRATIS employs mixture autoregressive (MAR) models to generate a new set of time series with diverse features. In statistical modelling, MAR models are commonly used to model populations with multiple statistical distributions and diverse characteristics, by using mixtures of models instead of a single autoregressive (AR) model. In the mixture of AR models, each AR process's coefficients are selected from a Gaussian distribution. Then, a mixture weight matrix provides the contribution of each AR model to the generated time series. For more detailed discussions of the GRATIS time series generation methodology, we refer to \cite{Kang2019-dy}. In our experiments, we use the implementation available in the \verb|generate_ts| function from the R package \verb|gratis| \citep{Kang2018-wm}.

In our experiments, we use GRATIS with the second approach (transfer strategy), which pre-trains a GFM model using the time series generated from GRATIS, and then transfers the knowledge to the target dataset. This is because the accuracy of GFMs can degenerate if the two joined time series datasets are too different from each other \citep{Bandara2019-iv}. Therefore, the augmented time series from the GRATIS method are presumably not suitable to be used with our first approach, the pooled strategy, which trains the original set of time series alongside the augmented time series. Using the second approach, even when using GRATIS, we can still transfer generic time series information from the augmented dataset. 

%\textbf{TODO: It would have still been worthwhile to run GRATIS.pooled nonetheless, just to check. But I guess it is too late now to do that.}

\subsection{Moving Block Bootstrapping}

%The boostrap aggregation, also known as bagging, is a technique that synthetically generates several versions of the training data to build multiple predictors, and averages over those predictors to obtain an aggregated predictor \citep{Breiman1996-el}. 
The MBB is a commonly used bootstrapping technique in time series forecasting \citep{Bergmeir2016-zk, Athanasopoulos2018-bn,Dantas2017-rv}.
In our experiments, we use the MBB technique for time series augmentation, following the procedure introduced in \cite{Bergmeir2016-zk}. To generate multiple copies of a time series, they first use STL to extract and subsequently remove seasonal and trend components of a time series. Next, the MBB technique is applied to the remainder of the time series, i.e., seasonally and trend adjusted series, to generate multiple versions of the residual components. For more detailed discussions of the MBB technique, we refer to \cite{Bergmeir2016-zk}. Finally, the bootstrapped residual components are added back together with the corresponding seasonal and trend to produce new bootstrapped versions of a time series. 
%In this way, the MBB technique can be used to generate new copies of a given time series with same trend and seasonality, but with different remainders. 
As original observations are used in the bootstrapping process, the artificially generated data closely resemble the distribution of the original training dataset, i.e., with similar seasonality and trend. We use the MBB implementation available in the \verb|bld.mbb.bootstrap| function from the R package \verb|forecast| \citep{Hyndman2015-vm}.

\subsection{Dynamic Time Warping Barycentric Averaging (DBA)}

Another procedure we use in our work is the Dynamic Time Warping (DTW) based time series augmentation technique proposed by \cite{Forestier2017-su}. In contrast to MBB that applies the bootstrapping procedure to each time series separately, the DBA approach averages a set of time series to generate new synthetic samples, thus being able to mix characteristics of different time series when generating new series, and therewith better accounting for the global characteristics in a group of time series. The DBA approach allows weighted averaging in the model when calculating the contribution of each time series towards the final generated time series. \cite{Forestier2017-su} develop three methods to determine the weights associated with the time series of the dataset, namely:  Average All (AA), Average Selected (AS), and Average Selected with Distance (ASD). For detailed discussions of the theoretical foundations and the methods, we refer to \cite{Forestier2017-su}. Having been specifically evaluated against time series classification tasks, DBA is used in this study in a time series forecasting setting. Following the competitive results shown in \cite{Forestier2017-su}, we use ASD as our primary averaging method. In particular, we use an implementation of the ASD method in Python from \cite{Petitjean2017-gp}. As characteristics of the original dataset are used to generate time series, similar to MBB, DBA can also be classified as a DA technique that generates augmented series similar to the original training dataset. 

\section{Experimental study}
\label{sec:experiments}
In this section, we evaluate the proposed variants of our framework on five time series datasets. First we describe the datasets, error metrics, statistical tests, hyper-parameter selection method, and benchmarks used in our experimental setup. Then, we provide a detailed analysis of the results obtained.

\subsection{Datasets}
\label{sec:datasets}
We use five benchmark time series datasets, which are composed of real-world applications and data from forecasting competitions. To limit the number of observations available and create a situation of relative data-scarcity, which is the main focus of our paper, for datasets with higher sampling rate, i.e., sub-hourly, hourly, and daily, we aggregate the series to weekly time series. %In this way, we constraint the amount of data available for training in our models, which allows us to evaluate our framework with less data-abundant settings. 
We briefly describe the five datasets as follows.

\begin{itemize}
\item NN5 Dataset~\citep{Crone2008-ye}: Daily dataset from the NN5 forecasting competition, containing daily cash withdrawals at various automatic teller machines (ATMs) located in the UK. We aggregated the original daily time series to weekly time series.
\item NN3 Dataset~\citep{Crone2011-vv}: Monthly dataset from the NN3 forecasting competition.
%\item Tourism Dataset~\citep{Athanasopoulos2011-ph}: Monthly dataset from the tourism forecasting competition.
\item AusEnergy-Demand Dataset ~\citep{Aemo2020-xx}: The energy demand of different states in Australia. The original time series has a data point every 15 minutes. We aggregate it to weekly energy demand.
\item AusGrid-Energy Dataset ~\citep{AusGrid2019-wq}: A collection of half-hourly time series, representing the energy consumption of households in Australia. We aggregate the original half-hourly time series to weekly time series.
\item Electricity Dataset ~\citep{Lai2018-yb}: Electricity consumption records, sampled every 15 minutes from multiple households in Portugal. We aggregate the original data to reflect weekly electricity consumption.
\end{itemize}

\begin{table}
\caption{Summary of the used datasets.}
\centering
\begin{tabular}{lcccccc}
	\toprule
	Dataset           		&N 			&$K_{min}$  &$K_{max}$ &T   		   &S    		&M  \\ \hline
	NN5   					&111	    &105		&105 		&weekly		   &52			&8	\\
	NN3   					&111		&51		    &126	    &monthly	   &12			&18	\\
	AusEnergy-Demand        &5			&313		&313		&weekly		   &52			&52 \\	
	AusGrid-Energy          &299		&132		&132		&weekly		   &52			&24 \\	
	Electricity           	&321		&146		&146		&weekly		   &52			&10 \\
	 \hline
\end{tabular}
\label{tab:datasummary}
\end{table}

Table \ref{tab:datasummary} summarises statistics of the datasets used in our experiments. Here, $N$ denotes the number of time series, $K_{min}$ and $K_{max}$ denote the minimum and maximum available lengths of the time series, respectively, $T$ denotes the sampling rate of the time series, $S$ represents the seasonality present in the time series, and $M$ is the intended forecast horizon. In the NN3 dataset, we see that the  lengths of the time series vary considerably, whereas other datasets contain time series with equal lengths ($K_{min}$ and $K_{max}$ are the same). Except for the NN3 dataset, we choose the size of the input window $n$ equivalent to $M*1.25$, following the heuristic proposed by \citet{Bandara2019-iv} and \citet{Hewamalage2019-il}. We use $n=11$ for the NN3 dataset due to the short lengths of its time series.

Furthermore, we choose the SE and DS training paradigms (see Section~\ref{sec:framework}) based on the recommendations of \cite{Bandara2020-zt}. We use the DS approach for the NN3 dataset, as the time series of those datasets are from disparate data sources, and have unknown starting dates. The SE approach is used for the remaining datasets, which are comprised of homogeneous time series with aligned time stamps.

\subsection{Performance Measures}
\label{sec:error}

To measure the performance of the proposed framework and benchmarks, we use two scale-independent evaluation metrics commonly found in the forecasting literature~\citep{Hyndman2006-ue}, namely the symmetric Mean Absolute Percentage Error (sMAPE) and the Mean Absolute Scaled Error (MASE). The sMAPE is defined as follows:
\begin{equation}
\text{sMAPE} = \frac{2}{m}\sum_{t=1}^{m}\left(\frac{\left|F_t - A_t\right|}{\left| F_t\right| + \left| A_t\right|} \right),
\label{smape}
\end{equation}
where $A_t$ represents the observation at time $t$, $F_t$ is the generated forecast, and $m$ indicates the forecast horizon. 
As the sMAPE can be unstable around zero values \citep{Hyndman2006-ue}, we use the modification proposed by \cite{Suilin2018-tc} for datasets that have values close to zero, namely the Electricity dataset in our benchmark suite. In this case, we modify the denominator of Equation~\eqref{smape} to $
{\max\left(\left| F_t\right| + \left| A_t\right| +\epsilon, 0.5 + \epsilon \right)}$, where $\epsilon$ is a small constant that we set to 0.1, following the recommendations of \cite{Suilin2018-tc}. 

%\begin{equation}
%\text{msMAPE} = \frac{2}{m}\sum_{t=1}^{m}\frac{\left|F_t - A_t\right|}{max%\left(\left| F_t\right| + \left| A_t\right| +\epsilon, 0.5 + \epsilon \right)}
%\label{msmape}
%\end{equation}

Moreover, we use the MASE, a less skewed and more interpretable error measure compared with sMAPE \citep{Hyndman2006-ue}. The MASE error measure is defined as follows:
\begin{equation}
\text{MASE} = \frac{\frac{1}{m}\sum_{t=1}^{m}|F_t - A_t|}{\frac{1}{n-S}\sum_{t=S+1}^{n}|A_{t}-A_{t-S}|}.
\label{mase}
\end{equation}
Additionally, in Equation~\eqref{mase}, $n$ is the number of observations in the training set of a time series, and $S$ refers to the length of the seasonal period in a given time series. The model evaluation of this study is presented using these error measures per series. We then calculate average ranks across all series from the benchmark suite, and also calculate Mean sMAPE, Median sMAPE, Mean MASE, and Median MASE across series within a dataset, to provide a broader overview of the error distributions.

\subsection{Statistical tests of the results}

We use the non-parametric Friedman rank-sum test to assess the statistically significance of differences among the compared forecasting methods on the benchmark datasets ~\citep{Garcia2010-jx}\footnote{More information can be found on the thematic web site of SCI2S about \emph{Statistical Inference in Computational Intelligence and Data Mining \url{http://sci2s.ugr.es/sicidm}}}. Also, Hochberg’s post-hoc procedure is used to further examine these differences with respect to the best performing technique. The statistical testing is done using the error measures specified in Section~\ref{sec:error}, with a significance level of $\alpha = 0.05$.

\subsection{Hyper-parameter Tuning and Data Augmentation}
\label{sec:hyperparameters}

The base learner of our forecast engine, LSTM, has various hyper-parameters, including LSTM cell dimension, number of epochs, hidden-layers, mini-batch size, and model regularisation terms. To autonomously determine the optimal values of these hyper-parameters, we use the sequential model-based algorithm configuration (SMAC), a variant of Bayesian Optimisation proposed by \cite{Hutter2011-wa}. In our experiments, we use the Python implementation of SMAC, which is available as a Python package \citep{AutoML_Group2017-bg}. To minimise the overall amount of hyper-parameters to be tuned in the learning phase, as the primary learning algorithm, we use COntinuous
COin Betting (COCOB) proposed by \cite{Orabona2017-ij}. Unlike in other gradient-based optimisation algorithms, such as Adam and Adagrad, COCOB does not require tuning of the learning rate. Instead, it attempts to minimise the loss function by self-tuning its learning rate. In this way, we remove the need for fine-tuning the learning rate of our optimisation algorithm. In our experiments, we use the Tensorflow implementation of COCOB \citep{Orabona2017-lj}. Table \ref{tab:parametergrid} summarises the ranges of hyper-parameter values explored in our experiments.

\begin{table}
\caption{The hyper-parameter ranges used in our experiments.}
\begin{center}
\begin{tabular}{lcc}
	\toprule
	Model Parameter            			&Minimum value 			&Maximum value\\ \hline
	LSTM cell dimension  				&20						&50	\\
	Mini-batch size						&1						&100\\
	Epoch size							&2						&5 \\
	Maximum epochs						&2						&50	\\
	Hidden layers						&1						&5\\
	Gaussian noise injection			&$10^{-4}$				&$8 \times 10^{-4}$\\
	Random-normal initialiser			&$10^{-4}$				&$8 \times 10^{-4}$\\
	L2-regularisation weight			&$10^{-4}$				&$8 \times 10^{-4}$\\ \hline
\end{tabular}
\label{tab:parametergrid}
\end{center}
\end{table}

The parameter uncertainty of our proposed models is addressed by training all the models on ten different Tensorflow graph seeds and taking the median of the forecasts generated with those seeds. When generating synthetic time series from the GRATIS, MBB, and DBA approaches (see Section~\ref{sec:dataaugmentation}), we use three different seeds to address the stochastic nature of these methods. We apply the proposed transfer and pooling strategies to each set of synthetic time series generated from those seeds and compute the average error across them. For each dataset, we generate an equal number of artificial time series from the proposed DA techniques. The number of generated time series ($n_{A}$) for each dataset is determined by the amount of bootstraps used per series in the MBB technique. In MBB, we use ten bootstraps per time series, except for the AusEnergy-Demand dataset, where we use 200 bootstraps per time series due to the small size of the dataset. For example, the $n_{A}$ values of the NN5, NN3, AusEnergy-Demand, AusGrid-Energy, and Electricity datasets are 1110, 1110, 1000, 2990, and 3210, respectively. Furthermore, when running the GRATIS method using the \verb|generate_ts| function, we set the $frequency$ parameter equal to the length of the seasonal period ($S$) of the time series. The $nComp$ parameter, which determines the number of mixing components in the MAR model, is set to 4, and $n$ is set to the maximum length of a series in the dataset ($K_{max}$).

%We also run GRATIS (see Section~\ref{sec:gratis}) using three different seeds to overcome the uncertainty in generating diverse groups of time series. The exact set of parameters used in the \verb|generate_ts| function are summarised in Table \ref{tab:gratisgrid}. Here, we set the $n.ts$ to 9000, as we expect to replace the source dataset $A$ (see Section~\ref{sec:modeltrainingplan}) from the time series generated by GRATIS. The $frequency$ parameter is 7, as the target dataset $B$ follows a weekly seasonality. According to \cite{Kang2019-dy}, the value of the $nComp$ can be ranged from (1-5). In our experiments, we set this value to 4, with the objective of compromising the time series diversity and the computational cost. The $n$ is equivalent to 550, as we use an identical time series length to our source dataset $A$.

%\begin{table}
%\caption{The model parameter settings used for time series generation using GRATIS}
%\begin{center}
%\begin{tabular}{lcc}
%	\toprule
%	Model Parameter            			&Description 									&Values\\ \hline
%	$n.ts$				  		&number of time series							&9000	\\
%	$frequency$					&seasonality of the time series					&7	\\
%	$nComp$						&number of mixing components in the MAR model	&4 \\
%	$n$							&length of the time series						&550\\ \hline
%\end{tabular}
%\label{tab:gratisgrid}
%\end{center}
%\end{table}

\subsection{Benchmarks and Variants}
\label{sec:benchmarks}

We use a host of univariate forecasting techniques to benchmark against our proposed GFM variants, including a forecasting method from the exponential smoothing family, namely the method ES as implemented in the \verb|smooth| package in R by \cite{Svetunkov2017-je}, and an ARIMA model from the \verb|forecast| package implemented in R \citep{Hyndman2015-vm}. As our exponential smoothing benchmark, we choose the ES implementation over the more popular ETS method from the \verb|forecast| package~\citep{Hyndman2008-yd}, as ES is not restricted by the number of seasonal coefficients to be included in the model. In addition to these well-established benchmarks from the time series forecasting literature, we use Prophet, a forecasting technique introduced by \cite{Taylor2017-lw}, in its implementation in the R package \verb|prophet|. 

Based on the different DA methods and TA architectures introduced in Sections~\ref{sec:transferlearning} and \ref{sec:dataaugmentation}, we define the following variants of our proposed framework.

\begin{description}
\item[LSTM.Baseline:] The baseline LSTM model that only uses the original set of time series to train a GFM.
\item[MBB.Pooled:] The LSTM model that uses the original set of time series pooled together with the synthetic time series generated from the MBB method to train a GFM.
\item[DBA.Pooled:] The LSTM model that uses the original set of time series pooled together with the synthetic time series generated from the DBA method to train a GFM.
\item[MBB.TL.$K$:] The LSTM model that transfers the knowledge using the TL architecture $K$, from a pre-trained LSTM model, which is trained across the time series generated from the MBB method.
\item[DBA.TL.$K$:] The LSTM model that transfers the knowledge using the TL architecture $K$, from a pre-trained LSTM model, which is trained across the time series generated from the DBA method.
\item[GRATIS.TL.$K$:] The LSTM model that transfers the knowledge using the TL architecture $K$, from a pre-trained LSTM model, which is trained across the time series generated from the GRATIS method.
\end{description}

\subsection{Computational Performance}

We report the computational costs in execution time of our proposed framework and the benchmark models on the NN5 dataset. The results on other datasets are comparable. The experiments are run on an Intel(R) i7 processor (3.2 GHz), with 2 threads per core, 6 cores in total, and 64GB of main memory.

\begin{table}[!tb]
\caption{The average ranking of each method across all the time series in the benchmark datasets, ordered by the first column, which is sMAPE. For each column, the results of the best performing method(s) are marked in boldface.}
\centering {%
\begin{tabular}{lrrrr}\hline
	 		Method           				&Rank sMAPE 	&Rank MASE\\ \hline
	 		DBA.TL.Dense.Freeze				&\textbf{10.60} &\textbf{10.63}\\
	 		MBB.TL.Dense.Retrain			&10.94			&10.95\\
	 		DBA.TL.LSTM.Freeze				&11.09			&11.14\\
	 		DBA.TL.Dense.Retrain			&11.17			&11.18\\
	 		DBA.TL.AddDense.Freeze			&11.23			&11.24\\
	 		DBA.Pooled						&11.39			&11.51\\
	 		DBA.TL.AddDense.Retrain			&11.52			&11.58\\
	 		GRATIS.TL.AddDense.Retrain		&11.55			&11.49\\
	 		GRATIS.TL.Dense.Retrain			&11.64			&11.59\\
	 		GRATIS.TL.LSTM.Retrain			&11.64			&11.66\\
	 		LSTM.Baseline					&11.74			&11.69\\
	 		MBB.TL.LSTM.Retrain				&11.87			&11.88\\	
	 		MBB.Pooled						&11.99         	&11.95\\
	 		DBA.TL.LSTM.Retrain				&12.08			&12.15\\
	 		GRATIS.TL.LSTM.Freeze			&12.16			&12.19\\
	 		MBB.TL.AddDense.Retrain			&12.37			&12.35\\
	 		MBB.TL.LSTM.Freeze				&13.05			&13.06\\
	 		MBB.TL.Dense.Freeze				&13.15			&13.14\\
	 		MBB.TL.AddDense.Freeze			&13.19			&13.10\\
	 		ARIMA							&14.22			&14.20\\
	 		Prophet 						&14.65			&14.49\\
	 		GRATIS.TL.Dense.Freeze			&14.95			&15.00\\
	 		GRATIS.TL.AddDense.Freeze		&14.99			&14.99\\	
	 		ES								&16.79			&16.82\\
			\hline
\end{tabular}
}
\label{tab:ranking}
\end{table}

\begin{table}[!tb]
\caption{Results of statistical testing for the sMAPE error measure across all the datasets. Adjusted p-values calculated from the Friedman test with Hochberg’s post-hoc procedure are shown. A horizontal line is used to separate the methods that perform significantly worse than the control method from the ones that do not.}
\centering
\begin{tabular}{llrrr}
	\toprule
			Method							&$p_{Hoch}$&\\ \hline
			DBA.TL.Dense.Freeze				&-\\
	 		MBB.TL.Dense.Retrain			&0.334\\
	 		DBA.TL.LSTM.Freeze				&0.317\\
	 		DBA.TL.Dense.Retrain			&0.300\\
	 		DBA.TL.AddDense.Freeze			&0.272\\
	 		DBA.Pooled						&0.110\\
	 		\hline
	 		DBA.TL.AddDense.Retrain			&0.047\\
	 		GRATIS.TL.AddDense.Retrain		&0.042\\
	 		GRATIS.TL.Dense.Retrain			&0.020\\
	 		GRATIS.TL.LSTM.Retrain			&0.020\\
	 		LSTM.Baseline					&0.009\\
	 		MBB.TL.LSTM.Retrain				&0.002\\	
	 		MBB.Pooled						&6.29 $\times$ $10^{-4}$\\
	 		DBA.TL.LSTM.Retrain				&2.18 $\times$ $10^{-4}$\\
	 		GRATIS.TL.LSTM.Freeze			&8.17 $\times$ $10^{-5}$\\
	 		MBB.TL.AddDense.Retrain			&4.04 $\times$ $10^{-6}$\\
	 		MBB.TL.LSTM.Freeze				&1.94 $\times$ $10^{-11}$\\
	 		MBB.TL.Dense.Freeze				&2.21 $\times$ $10^{-12}$\\
	 		MBB.TL.AddDense.Freeze			&9.24 $\times$ $10^{-13}$\\
	 		ARIMA							&1.22 $\times$ $10^{-24}$\\
	 		Prophet 						&1.11 $\times$ $10^{-30}$\\	
	 		GRATIS.TL.Dense.Freeze			&2.30 $\times$ $10^{-35}$\\
	 		GRATIS.TL.AddDense.Freeze		&6.91 $\times$ $10^{-36}$\\
	 		ES								&5.10 $\times$ $10^{-71}$\\	
	\hline			
\end{tabular}
\label{tab:smapeallstat}
\end{table}

\begin{table}[!tb]
\caption{Results of statistical testing for the MASE error measure across all the datasets. Adjusted p-values calculated from the Friedman test with Hochberg’s post-hoc procedure are shown. A horizontal line is used to separate the methods that perform significantly worse than the control method from the ones that do not.}
\centering
\begin{tabular}{llrrr}
	\toprule
			Method							&$p_{Hoch}$&\\ \hline
			DBA.TL.Dense.Freeze				&-\\
	 		MBB.TL.Dense.Retrain			&0.396\\
	 		DBA.TL.LSTM.Freeze				&0.396\\
	 		DBA.TL.Dense.Retrain			&0.396\\
	 		DBA.TL.AddDense.Freeze			&0.396\\
	 		DBA.TL.AddDense.Retrain			&0.396\\
	 		GRATIS.TL.Dense.Retrain			&0.396\\
	 		GRATIS.TL.AddDense.Retrain		&0.396\\
	 		LSTM.Baseline					&0.191\\
	 		GRATIS.TL.LSTM.Retrain			&0.191\\
	 		DBA.Pooled						&0.161\\
	 		\hline	 		
	 		MBB.TL.LSTM.Retrain				&0.036\\	
	 		MBB.Pooled						&0.001\\
	 		GRATIS.TL.LSTM.Freeze			&1.11 $\times$ $10^{-4}$\\
	 		DBA.TL.LSTM.Retrain				&2.11 $\times$ $10^{-5}$\\
	 		MBB.TL.AddDense.Retrain			&1.26 $\times$ $10^{-6}$\\
	 		MBB.TL.LSTM.Freeze				&6.36 $\times$ $10^{-11}$\\
	 		MBB.TL.Dense.Freeze				&3.30 $\times$ $10^{-13}$\\
	 		MBB.TL.AddDense.Freeze			&1.06 $\times$ $10^{-13}$\\
	 		ARIMA							&3.45 $\times$ $10^{-26}$\\
	 		GRATIS.TL.Dense.Freeze			&1.71 $\times$ $10^{-29}$\\
	 		GRATIS.TL.AddDense.Freeze		&4.34 $\times$ $10^{-30}$\\
	 		Prophet 						&1.27 $\times$ $10^{-30}$\\	
	 		ES								&1.13 $\times$ $10^{-71}$\\	
	\hline			
\end{tabular}
\label{tab:maseallstat}
\end{table}

\begin{table}[!tb]
\caption{The Mean sMAPE results across all the benchmark datasets. For each dataset, the results of the best performing method(s) are marked in boldface.}
\centering
\resizebox{\columnwidth}{!}{%
\begin{tabular}{lccccc}\hline
	 		Method      			&AusEnergy-Demand	&AusGrid-Energy  &Electricity 	&NN3 	    &NN5\\ \hline
	 		LSTM.Baseline			&0.0550				&0.2235			 &0.0900		&0.1655		&0.1075\\
	 		MBB.Pooled				&\textbf{0.0518}	&0.2228			 &0.0903		&0.1682		&0.1080\\
	 		DBA.Pooled				&0.0607				&0.2178			 &0.0884		&0.1650		&0.1071\\
	 		MBB.TL.Dense.Freeze		&0.0539			 	&0.2324			 &0.0890        &0.1659		&0.1169\\
	 		MBB.TL.Dense.Retrain	&0.0558				&0.2195			 &0.0884		&0.1661		&0.1068\\
	 		MBB.TL.AddDense.Freeze	&0.0539				&0.2325			 &0.0893		&0.1665		&0.1166\\
	 		MBB.TL.AddDense.Retrain	&0.0545				&0.2214			 &0.0900		&0.1667		&0.1090\\
	 		MBB.TL.LSTM.Freeze		&0.0548				&0.2275			 &0.0928		&0.1658		&0.1154\\
	 		MBB.TL.LSTM.Retrain		&0.0556				&0.2248			 &0.0887		&0.1663		&0.1086\\
			DBA.TL.Dense.Freeze		&0.0561				&0.2150			 &0.0867		&0.1654		&0.1185\\
			DBA.TL.Dense.Retrain	&0.0565				&0.2172			 &0.0881		&0.1661		&0.1072\\	 		
	 		DBA.TL.LSTM.Freeze		&0.0560				&0.2187			 &0.0869		&0.1653		&0.1166\\
	 		DBA.TL.LSTM.Retrain		&0.0551				&0.2225			 &0.0896		&0.1639		&0.1132\\
	 		DBA.TL.AddDense.Freeze	&0.0539				&\textbf{0.2147} &0.0871		&0.1660		&0.1242\\
	 		DBA.TL.AddDense.Retrain	&0.0553				&0.2175			 &0.0882		&0.1664		&0.1073\\
	 		GRATIS.TL.Dense.Freeze	&0.0639				&0.2629			 &\textbf{0.0853}&0.1673	&0.1245\\
	 		GRATIS.TL.Dense.Retrain	&0.0543				&0.2233			 &0.0870		&0.1664		&0.1067\\
	 		GRATIS.TL.AddDense.Freeze &0.0643			&0.2607			 &\textbf{0.0853}&0.1675	&0.1264\\
	 		GRATIS.TL.AddDense.Retrain&0.0597			&0.2221			 &0.0867		&0.1669		&\textbf{0.1066}\\
			GRATIS.TL.LSTM.Freeze	&0.0566				&0.2261			 &0.0865		&0.1659		&0.1189\\
			GRATIS.TL.LSTM.Retrain	&0.0557				&0.2216			 &0.0879		&0.1648		&0.1090\\
	 		Prophet					&0.0595				&0.2566			 &0.0996		&0.2518		&0.1143\\
			ES						&0.0642				&0.3318			 &0.1136		&\textbf{0.1532}&0.1211\\    
	 		ARIMA					&0.0739				&0.2619			 &0.0974		&0.1564		&0.1355\\
			\hline
\end{tabular}
}
\label{tab:meansMAPE}
\end{table}

\begin{table}[!tb]
\caption{The Median sMAPE results across all the benchmark datasets. For each dataset, the results of the best performing method(s) are marked in boldface.}
\centering
\resizebox{\columnwidth}{!}{%
\begin{tabular}{lccccc}\hline
	 		Method      			&AusEnergy-Demand	&AusGrid-Energy  &Electricity 	&NN3 	    &NN5\\ \hline
	 		LSTM.Baseline			&0.0536				&0.1949			 &0.0590		&0.1170		&0.1028\\
	 		MBB.Pooled				&\textbf{0.0504}	&0.1933			 &0.0583		&0.1187		&0.1014\\
	 		DBA.Pooled				&0.0590				&0.1858			 &0.0618		&0.1174		&\textbf{0.1013}\\
	 		MBB.TL.Dense.Freeze		&0.0532				&0.1998			 &\textbf{0.0563}&0.1172	&0.1088\\
			MBB.TL.Dense.Retrain	&0.0540				&0.1880			 &0.0574		&0.1165		&0.1034\\	 		
	 		MBB.TL.AddDense.Freeze	&0.0533				&0.1966			 &0.0564		&0.1170		&0.1073\\
			MBB.TL.AddDense.Retrain	&0.0542				&0.1896			 &0.0570		&0.1180		&0.1040\\
			MBB.TL.LSTM.Freeze		&0.0536				&0.1934			 &0.0592		&0.1175		&0.1090\\
	 		MBB.TL.LSTM.Retrain		&0.0538				&0.1930			 &0.0571		&0.1171		&0.1018\\
	 		DBA.TL.Dense.Freeze		&0.0545				&\textbf{0.1836} &0.0571		&0.1173		&0.1067\\
	 		DBA.TL.Dense.Retrain	&0.0540				&0.1859			 &0.0587		&0.1172		&0.1022\\	 		
	 		DBA.TL.AddDense.Freeze	&0.0548				&0.1853			 &0.0573		&0.1172		&0.1100\\
	 		DBA.TL.AddDense.Retrain	&0.0540				&0.1870			 &0.0586		&0.1174		&0.1052\\ 
	 		DBA.TL.LSTM.Freeze		&0.0545				&0.1865			 &0.0578		&0.1170		&0.1065\\
	 		DBA.TL.LSTM.Retrain		&0.0544				&0.1950			 &0.0603		&0.1177		&0.1059\\
	 		GRATIS.TL.Dense.Freeze	&0.0631				&0.2389			 &0.0572		&0.1174		&0.1085\\
	 		GRATIS.TL.Dense.Retrain	&0.0534				&0.1929			 &0.0566		&0.1185		&0.1039\\
	 		GRATIS.TL.AddDense.Freeze&0.0643			&0.2371			 &0.0576		&0.1176		&0.1089\\
	 		GRATIS.TL.AddDense.Retrain&0.0574			&0.1892			 &0.0564		&0.1176		&0.1045\\
	 		GRATIS.TL.LSTM.Freeze	&0.0578				&0.1926			 &0.0569		&0.1172		&0.1081\\
	 		GRATIS.TL.LSTM.Retrain	&0.0548				&0.1895			 &0.0577		&0.1166		&0.1028\\	 
	 		Prophet					&0.0565				&0.2158			 &0.0621		&0.1926		&0.1062\\
			ARIMA					&0.0580				&0.2216			 &0.0660		&0.1182		&0.1220\\
			ES						&0.0606				&0.2852			 &0.0934		&\textbf{0.1135}&0.1079\\

			\hline
\end{tabular}
} 
\label{tab:mediansMAPE}
\end{table}

\begin{table}[!tb]
\caption{The Mean MASE results across all the benchmark datasets. For each dataset, the results of the best performing method(s) are marked in boldface.}
\centering
\resizebox{\columnwidth}{!}{%
\begin{tabular}{lccccc}\hline
	 		Method      			&AusEnergy-Demand	&AusGrid-Energy  &Electricity 	&NN3 	    &NN5\\ \hline
	 		LSTM.Baseline			&0.9564				&0.8183			 &0.7512		&0.9471		&0.7999\\
	 		MBB.Pooled				&\textbf{0.9073}	&0.8271			 &0.7497		&0.9625		&0.8050\\
	 		DBA.Pooled				&1.0504				&0.8087			 &0.7741		&0.9459		&0.7946\\
	 		MBB.TL.Dense.Freeze		&0.9372				&0.8661			 &0.7475		&0.9494		&0.8602\\
			MBB.TL.Dense.Retrain	&0.9718				&0.8117			 &0.7385		&0.9499		&0.7942\\	 		
	 		MBB.TL.AddDense.Freeze	&0.9357				&0.8686			 &0.7483		&0.9518		&0.8576\\
			MBB.TL.AddDense.Retrain	&0.9444				&0.8233			 &0.7577		&0.9580		&0.8097\\
			MBB.TL.LSTM.Freeze		&0.9530				&0.8483			 &0.7690		&0.9509		&0.8485\\
	 		MBB.TL.LSTM.Retrain		&0.9690				&0.8396			 &0.7338		&0.9533		&0.8094\\
	 		DBA.TL.Dense.Freeze		&0.9669				&0.7879			 &0.7393		&0.9463		&0.8733\\
	 		DBA.TL.Dense.Retrain	&0.9858				&0.7990			 &0.7523		&0.9501		&0.7972\\	 		
	 		DBA.TL.AddDense.Freeze	&0.9234				&\textbf{0.7874} &0.7438		&0.9496		&0.9142\\
	 		DBA.TL.AddDense.Retrain	&0.9684				&0.8008			 &0.7517		&0.9547		&0.7981\\ 
	 		DBA.TL.LSTM.Freeze		&0.9717				&0.8036			 &0.7321		&0.9492		&0.8581\\
	 		DBA.TL.LSTM.Retrain		&0.9626				&0.8229			 &0.7601		&0.9371		&0.8396\\
	 		GRATIS.TL.Dense.Freeze	&1.1032				&0.9974			 &0.7378		&0.9572		&0.9142\\
	 		GRATIS.TL.Dense.Retrain	&0.9402				&0.8242			 &0.7306		&0.9550		&0.7929\\
	 		GRATIS.TL.AddDense.Freeze &1.0921			&0.9859			 &0.7375		&0.9582		&0.9280\\
	 		GRATIS.TL.AddDense.Retrain &1.0235			&0.8200			 &0.7306		&0.9567		&\textbf{0.7921}\\
	 		GRATIS.TL.LSTM.Freeze	&0.9676				&0.8383			 &\textbf{0.7305}&0.9509	&0.8732\\
	 		GRATIS.TL.LSTM.Retrain	&0.9635				&0.8148			 &0.7415		&0.9443		&0.8104\\	 
	 		Prophet					&1.0627				&0.9039			 &0.8034		&1.3233		&0.8496\\
			ARIMA					&1.2111				&0.9603			 &0.8379		&0.9235		&1.0021\\
			ES						&1.0473				&1.2839			 &1.1696		&\textbf{0.8942}&0.8918\\
			\hline
\end{tabular}
} 
\label{tab:meanMASE}
\end{table}

\begin{table}[!tb]
\caption{The Median MASE results across all the benchmark datasets. For each dataset, the results of the best performing method(s) are marked in boldface.}
\centering
\resizebox{\columnwidth}{!}{%
\begin{tabular}{lccccc}\hline
	 		Method      			&AusEnergy-Demand	&AusGrid-Energy  &Electricity 	&NN3 	    &NN5\\ \hline
	 		LSTM.Baseline			&0.8552				&0.7508			 &0.6982		&0.7518		&0.7374\\
	 		MBB.Pooled				&0.8575				&0.7399			 &0.6951		&0.7730		&0.7579\\
	 		DBA.Pooled				&0.9926				&0.7210			 &0.7097		&0.7684		&0.7340\\
	 		MBB.TL.Dense.Freeze		&0.8638				&0.8036			 &0.6919		&0.7806		&0.7751\\
			MBB.TL.Dense.Retrain	&0.8796				&0.7427			 &0.6731		&0.7629		&0.7342\\	 		
	 		MBB.TL.AddDense.Freeze	&\textbf{0.8499}	&0.802			 &0.6850		&0.7629		&0.7597\\
			MBB.TL.AddDense.Retrain	&0.8671				&0.7574			 &0.7145		&0.7735		&0.7458\\
			MBB.TL.LSTM.Freeze		&0.8693				&0.7726			 &0.7118		&0.7519		&0.7464\\
	 		MBB.TL.LSTM.Retrain		&0.8710				&0.7640			 &0.6859		&0.7655		&0.7201\\
	 		DBA.TL.Dense.Freeze		&0.8721				&\textbf{0.7130} &\textbf{0.6558}&0.7735	&0.7782\\
	 		DBA.TL.Dense.Retrain	&0.8750				&0.7308			 &0.6938		&0.7653		&\textbf{0.7186}\\
	 		DBA.TL.AddDense.Freeze	&0.8516				&0.7140			 &0.6600		&0.7551		&0.8284\\
	 		DBA.TL.AddDense.Retrain	&0.8697				&0.7287			 &0.6976		&0.7734		&0.7451\\ 
	 		DBA.TL.LSTM.Freeze		&0.8818				&0.7464			 &0.6574		&0.7725		&0.7534\\
	 		DBA.TL.LSTM.Retrain		&0.8720				&0.7601			 &0.6965		&\textbf{0.7438}&0.7720\\
	 		GRATIS.TL.Dense.Freeze	&1.1804				&0.9192			 &0.6695		&0.7705		&0.8415\\
	 		GRATIS.TL.Dense.Retrain	&0.8565				&0.7430			 &0.6821		&0.7696		&0.7529\\
	 		GRATIS.TL.AddDense.Freeze &1.0672			&0.9122			 &0.6630		&0.7675		&0.8548\\
	 		GRATIS.TL.AddDense.Retrain &0.9173		   	&0.7431			 &0.6824		&0.7590		&0.7322\\
	 		GRATIS.TL.LSTM.Freeze	&0.8951				&0.7628			 &0.6779		&0.7491		&0.8055\\
	 		GRATIS.TL.LSTM.Retrain	&0.8703				&0.7516			 &0.6745		&0.7471		&0.7398\\	 
	 		Prophet					&0.9809				&0.8394			 &0.7245		&1.2338		&0.8625\\
			ARIMA					&1.0978				&0.8300			 &0.7884		&0.7828		&0.9385\\
			ES						&1.0652				&1.0525			 &1.0181		&0.7793		&0.7620\\
			\hline
\end{tabular}
} 
\label{tab:medianMASE}
\end{table}

\subsection{Results and Discussion}

Table~\ref{tab:ranking} summarises the overall performance of the proposed variants and the benchmarks, in terms of average ranking across all series in the benchmark suite. According to Table~\ref{tab:ranking}, the proposed DBA.TL.Dense.Freeze variant obtains the best Rank sMAPE and Rank MASE. It can be seen that many of the proposed variants outperform the baseline model, LSTM.Baseline on both evaluation metrics. Furthermore, except for the GRATIS.TL.Dense.Freeze and GRATIS.TL.AddDense.Freeze variants, all other proposed methods obtain better accuracies than the benchmarks ES, ARIMA, and Prophet.

Regarding statistical testing, the overall result of the Friedman rank sum test for sMAPE is a $p$-value of $2.66 \times 10^{-10}$, which means the results are highly significant.
Table~\ref{tab:smapeallstat} shows the results of the post-hoc test. The DBA.TL.Dense.Freeze method performs best and is chosen as the control method. We can see from the table that the improvements in accuracy over the baseline LSTM.Baseline and the benchmarks are highly significant.

Table~\ref{tab:maseallstat} shows the results of the statistical testing evaluation for the MASE error measure. The overall result of the Friedman rank sum test for MASE is a $p$-value of $2.58 \times 10^{-10}$, which means the results are highly significant. The DBA.TL.Dense.Freeze variant performs best and is again chosen as the control method. We see that the improvements over the LSTM.Baseline variant are not statistically significant in this instance, but the improvements over the benchmarks ES, Prophet, and ARIMA are highly significant. 

After the analysis of the results overall across datasets, where we have been able to establish the statistical significance of the accuracy gains of our methods, we further investigate error distributions in more detail for each dataset. 
The results of all the proposed variants in terms of the mean sMAPE metric are shown in Table~\ref{tab:meansMAPE}. 
We see that MBB.Pooled, DBA.TL.AddDen\-s\-e.Retrain, GRATIS.TL.Dense.Freeze, GRA\-T\-IS.TL.AddDense.Freeze, and  GRATIS.TL.AddDense.Retrain obtain the best Mean sMAPE for the AusEnergy-Demand, AusGrid-Energy, Electricity, and NN5 datasets. Whereas, the ES method achieves the best Mean sMAPE for the NN3 dataset. 
For the AusEnergy-Demand dataset, we observe that the majority of the MBB based knowledge transfer variants (both based on TL and pooled), can achieve better accuracy than the LSTM.Baseline. 
In terms of the AusGrid-Energy dataset, it can be seen that, in most cases, the DBA-based knowledge transfer variants obtain better forecasts than LSTM.Baseline, which is contrary to our previous findings from the AusEnergy-Demand dataset. It is also noteworthy to mention that both pooled variants, i.e., DBA.Pooled and MBB.Pooled, outperform the LSTM.Baseline in this dataset. 
For the Electricity dataset, we observe that all the GRATIS and DBA based variants outperform the LSTM.Baseline benchmark. Also, among the pooled variants, we see that only the DBA.Pooled method performs better than LSTM.Baseline. Moreover, it can be seen that all the variants that use TL.Dense.Freeze, TL.Dense.Retrain, TL.AddDense.Freeze, TL.AddDense.Retrain, and TL.LSTM.Retrain as the TL architecture, generate better forecasts than the baseline. 
For the NN3 dataset, even though our proposed variants are unable to outperform the statistical benchmarks, we see that some of the DBA based variants outperform our baseline model in terms of Mean sMAPE. Finally, with respect to the NN5 dataset, on average, we see that the variants that use the TL.Dense.Retrain and TL.AddDense.Retrain architectures achieve better accuracies compared with variants that use other TL architectures.
Also, except for the NN3 dataset, we see that many of the proposed variants outperform the state-of-the-art forecasting methods, such as ES, Prophet, and ARIMA variants on Mean sMAPE.

Table~\ref{tab:mediansMAPE} shows the results of all the proposed variants in terms of the median sMAPE metric. It can be seen that the proposed  MBB.Pooled method, DBA.TL.Dense.Freeze method, MBB.TL.Dense.Freeze and DBA.Pooled method obtain the best Median sMAPE for the AusEnergy-Demand, AusGrid-Energy, Electricity, and NN5 datasets, respectively. The ES method achieves the best Median sMAPE for the NN3 dataset.
Similar to the previous findings from Table~\ref{tab:meansMAPE}, we see that for the AusEnergy-Demand dataset, the majority of the MBB based knowledge transfer variants generate better forecasts compared with the LSTM.Baseline. 
Also, for the AusGrid-Energy dataset, the results indicate that the DBA based knowledge transfer variants obtain better accuracy than the MBB based knowledge transfer variants and the LSTM.Baseline. Furthermore, both DBA.Pooled and the MBB.Pooled variants outperform the LSTM.Baseline for this dataset. 
In terms of the Electricity dataset, it can be seen that except for the DBA.Pooled, MBB.TL.LSTM.Freeze, and DBA.TL.LSTM.Retrain methods, and all other proposed variants manage to outperform the LSTM.Baseline benchmark. Moreover, among the pooled variants, it can be seen that only the MBA.Pooled method performs better than the LSTM.Baseline.
For the NN3 dataset, we see that only the DBA.TL.LSTM.Freeze and GRATIS.TL.LSTM.Retrain variants can generate better accuracies compared with our baseline model. Concerning the NN5 dataset, on average, we see that both pooled variants can outperform the LSTM.Baseline benchmark. Overall, in terms of the Mean sMAPE error measure, we notice that the majority of our proposed methods achieve better results than the statistical benchmarks.

The results of the proposed variants in terms of the mean MASE error metric are as shown in Table~\ref{tab:meanMASE}. Apart from the NN3 dataset, we note that our proposed variants obtain the best accuracies for the rest of the datasets. In terms of the AusEnergy-Demand dataset, we observe that the MBB.Pooled variant achieves the best results outperforming the LSTM.Baseline.
We note that the DBA.TL.AddDense.Freeze method obtains the best results for the AusGrid-Energy dataset. Here, we see that the DBA.Pooled method and the majority of the DBA based TL architectures achieve better results compared with our baseline variant. 
In terms of the Electricity dataset, we observe that the proposed GRATIS.TL.LSTM.Freeze achieves the best Mean MASE. Moreover, it can be seen that all the GRATIS based TL architectures outperform the LSTM.Baseline.
Similar to the previous observations from Table~\ref{tab:meansMAPE} and Table~\ref{tab:mediansMAPE}, we see that the ES benchmark obtains the best Mean MASE for the NN3 dataset. However, we note that the DBA.Pooled method, DBA.TL.Dense.Freeze method, DBA.TL.LSTM.Retrain method and GRATIS.TL.LSTM.Retrain method outperform the LSTM.Baseline with respect to the Mean MASE.
The GRATIS.TL.AddDense.Retrain variant achieves the best Mean MASE for the NN5 dataset. In addition to the GRATIS.TL.AddDense.Retrain method, and we see that the proposed DBA.Pooled method, MBB.TL.Dense.Retrain method, DBA.TL.Dense.Retrain method, DBA.TL.AddDense.Retrain method, and GRATIS.TL.Dense.Retrain method obtain better Mean MASE values compared with the LSTM.Baseline.
Overall, with respect to the Mean MASE error measure, we see that in the majority of cases, our proposed methods outperform the statistical benchmarks.

Table~\ref{tab:medianMASE} shows the results of all the proposed variants in terms of the median MASE metric. We see that the proposed MBB.TL.AddDense.Freeze variant obtains the best Median MASE for the AusEnergy-Demand dataset. 
Whereas, the DBA.TL.Dense.Freeze method obtains the best results for both AusGrid-Energy and Electricity datasets. For the AusGrid-Energy dataset, it can be seen that both the MBB.Pooled and DBA.Pooled methods outperform the LSTM.Baseline. Also, the majority of the DBA based TL architectures achieve better results compared with our baseline variant. 
For the Electricity dataset, it can be seen that DBA and GRATIS based TL architectures generate accurate forecasts compared with the LSTM.Baseline.
The DBA.TL.LSTM.Retrain variant achieves the best Median MASE for the NN3 dataset. This is contrary to our previous findings from Table~\ref{tab:meansMAPE}, Table~\ref{tab:mediansMAPE}, and Table~\ref{tab:meanMASE}, in which the statistical benchmarks outperformed our variants on the NN3 dataset. Also, we notice that the majority of the proposed variants achieve better Median MASE values compared with the Prophet, ARIMA, and ES benchmarks.
In terms of the NN5 dataset, we see that the proposed DBA.TL.Dense.Retrain method achieves the best results. Furthermore, we see that the proposed DBA.Pooled method, MBB.TL.Dense.Retrain method, MBB.TL.LSTM.Retrain method, and GRATS.TL.AddDe\-n\-se.Retrain method outperform the LSTM.Baseline.
Similar to our previous findings from Table~\ref{tab:meansMAPE}, Table~\ref{tab:mediansMAPE}, and Table~\ref{tab:meanMASE}, we see that the majority of our proposed variants outperform the statistical benchmarks concerning the Median MASE error measure.

\begin{table}[!tb]
\caption{The computational time for the NN5 Dataset, ordered by the last column, which is the total computation time in seconds.}
\centering
\begin{tabular}{lrrr}\hline
	 		Method           				&Timeseries-generation	&Model-Training  &Total-time \\ \hline
	 		ARIMA							&-						&19			 	 &19\\
	 		Prophet 						&-						&40			 	 &40\\	
	 		ES								&-						&120 		 	 &120\\
	 		LSTM.Baseline					&-						&994			 &994\\
	 		MBB.Pooled						&4         				&1253          	 &1257\\
	 		DBA.Pooled						&13						&1253			 &1266\\
			DBA.TL.AddDense.Freeze			&13						&1629			 &1642\\	 		
	 		MBB.TL.AddDense.Freeze			&4						&1675			 &1679\\
	 		DBA.TL.Dense.Freeze				&13	        			&1848			 &1861\\
	 		MBB.TL.Dense.Freeze				&4						&1970			 &1974\\ 
			GRATIS.TL.Dense.Freeze			&160					&2167		 	 &2327\\
			GRATIS.TL.AddDense.Freeze		&160					&2188			 &2348\\	 		
	 		DBA.TL.AddDense.Retrain			&13						&2747			 &2760\\
	 		MBB.TL.Dense.Retrain			&4						&3057			 &3061\\
	 		MBB.TL.AddDense.Retrain			&4						&3060			 &3064\\
	 		DBA.TL.Dense.Retrain			&13						&3732			 &3745\\
	 		GRATIS.TL.AddDense.Retrain		&160					&4560			 &4720\\
	 		DBA.TL.LSTM.Freeze				&13						&4454			 &4467\\
	 		MBB.TL.LSTM.Freeze				&4						&4623			 &4627\\
	 		DBA.TL.LSTM.Retrain				&13						&4575			 &4588\\
	 		GRATIS.TL.Dense.Retrain			&160					&4575			 &4735\\
	 		MBB.TL.LSTM.Retrain				&4						&5209			 &5213\\
	 		GRATIS.TL.LSTM.Freeze			&160					&5510			 &5670\\
	 		LGRATIS.TL.LSTM.Retrain			&160					&6283			 &6443\\
			\hline
\end{tabular}
\label{tab:time}
\end{table}

Table~\ref{tab:time} provides a summary of computational cost of the proposed variants and benchmarks on the NN5 dataset. According to Table~\ref{tab:time}, we see that the statistical benchmarks, such as ARIMA, Prophet and ES, have a lower execution time compared with the proposed variants. Nonetheless, according to Table~\ref{tab:meansMAPE}, Table~\ref{tab:mediansMAPE}, Table~\ref{tab:meanMASE}, Table~\ref{tab:medianMASE}, we see that these benchmarks do not display competitive results compared with the proposed variants. With respect to computational cost among the proposed methods, we observe that the variants that use pooled approach for knowledge transfer have less computational time than TL-based methods. This is due to their additional model pre-training and model transfer procedure. Furthermore, as the complexity of the TL architecture increases, the total computational time also gradually increases. Here, the TL based variants that freeze the initial layers and do not re-train the layers of the pre-trained model, i.e., TL.Dense.Freeze, TL.AddDense.Freeze, and TL.LSTM.Freeze, consume a smaller amount of time than their counterpart architectures that re-train the layers of the pre-trained model, i.e., TL.Dense.Retrain, TL.AddDense.Retrain, and MBB.TL.LSTM.Retrain (see Section~\ref{sec:transferlearning}). Also, with respect to the time series data augmentation techniques, we see that the GRATIS method takes the highest amount of time to generate the target time series.

To summarise, the proposed DBA based variants achieve competitive results in our experiments. One exception to this is the AusEnergy-Demand dataset, where MBB based variants outperform the DBA based methods. 
It can be mainly attributed to the small size of the AusEnergy-Demand dataset, where the DBA technique performs poorly as the number of time series in the source dataset is limited. As the MBB technique generates time series independent of the number of time series available in the source dataset as it augments each time series in isolation, the MBB based variants outperform the DBA based variants in this scenario. The better performance of DBA and MBB based variants over GRATIS based variants also indicates that DA techniques that generate time series similar to the original distribution of the dataset contribute more towards improving the base accuracy of the models. The results also indicate that, in most cases, the proposed variants can outperform state-of-the-art statistical forecasting techniques, such as ES, ARIMA, and Prophet. In situations where our methods are unable to perform better than the statistical benchmarks, we observe that the majority of our proposed variants can improve the accuracy of the baseline model.

\section{Conclusions}
\label{sec:con}
Generating accurate forecasts with a limited number of time series can be a challenging task for global forecasting models that train across all the available time series. In this study, we have introduced a novel data augmentation based forecasting framework to supplement recurrent neural network based global model architectures, when used in settings with limited amounts of available data. We have used three time series augmentation techniques to produce time series synthetically. They include the Moving Block Bootstrap and the Dynamic Time-Warping Barycentric Averaging techniques that are capable of generating time series that are similar to those in the original dataset, and the GRATIS method that generates time series with diverse characteristics, which can be dissimilar to those in the original dataset.

To transfer knowledge representations from the augmented dataset to the target dataset with less data, we have employed two strategies; the pooled and transfer learning strategies. The pooled strategy trains on the augmented time series together with the original time series database, while the transfer learning strategy initially pre-trains a global model using the augmented time series, and then transfers the pre-trained knowledge representations to the target dataset using various transfer learning architectures.

We have evaluated our methods using five benchmark datasets, including two competition datasets and three real-world datasets. The results have shown that the proposed variants achieve competitive results under small to medium training set size conditions, outperforming the baseline global model and many state-of-the-art univariate forecasting methods with statistical significance. The results also indicate that data augmentation techniques that generate time series with similar characteristics to the target dataset achieve better results than those that generate time series with diverse characteristics. Nonetheless, the results suggest that the subset of GRATIS based variants, which re-trains the pre-trained model and newly added layers, can be a competitive approach among the baseline global model and univariate forecasting methods. This highlights the fact that resembling the general characteristics of time series, and then transferring this information to the target dataset can be useful to improve model accuracy, even if the augmented time series are diverse and different from the target dataset. Furthermore, we observe that the choice of the proposed strategies can be determined by the size of the original dataset, where the pooling strategy is more suitable for situations where the size of the original dataset is small, and the transfer strategy is better if the dataset is more extensive.

As a possible future work, more sophisticated feature extraction techniques, such as Encoder-Decoder architectures could be used, replacing the stacking architecture of our model. The extracted, latent features could then be re-purposed to train on a target dataset for forecasting.

\section*{Acknowledgements}
This research was supported by the Australian Research Council under grant DE190100045, by a Facebook Statistics for Improving Insights and Decisions research award, by Monash Institute of Medical Engineering seed funding, by the MASSIVE - High performance computing facility, Australia, and by the National Natural Science Foundation of China
(No. 11701022).

\bibliographystyle{elsarticle-harv}
\bibliography{reference}

\begin{thebibliography}{86}
\expandafter\ifx\csname natexlab\endcsname\relax\def\natexlab#1{#1}\fi
\providecommand{\url}[1]{\texttt{#1}}
\providecommand{\href}[2]{#2}
\providecommand{\path}[1]{#1}
\providecommand{\DOIprefix}{doi:}
\providecommand{\ArXivprefix}{arXiv:}
\providecommand{\URLprefix}{URL: }
\providecommand{\Pubmedprefix}{pmid:}
\providecommand{\doi}[1]{\href{http://dx.doi.org/#1}{\path{#1}}}
\providecommand{\Pubmed}[1]{\href{pmid:#1}{\path{#1}}}
\providecommand{\bibinfo}[2]{#2}
\ifx\xfnm\relax \def\xfnm[#1]{\unskip,\space#1}\fi
%Type = Article
\bibitem[{Abadi et~al.(2016)Abadi, Agarwal, Barham, Brevdo, Chen, Citro,
  Corrado, Davis, Dean, Devin, Ghemawat, Goodfellow, Harp, Irving, Isard, Jia,
  Jozefowicz, Kaiser, Kudlur, Levenberg, Mane, Monga, Moore, Murray, Olah,
  Schuster, Shlens, Steiner, Sutskever, Talwar, Tucker, Vanhoucke, Vasudevan,
  Viegas, Vinyals, Warden, Wattenberg, Wicke, Yu and Zheng}]{Abadi2016-rr}
\bibinfo{author}{Abadi, M.}, \bibinfo{author}{Agarwal, A.},
  \bibinfo{author}{Barham, P.}, \bibinfo{author}{Brevdo, E.},
  \bibinfo{author}{Chen, Z.}, \bibinfo{author}{Citro, C.},
  \bibinfo{author}{Corrado, G.S.}, \bibinfo{author}{Davis, A.},
  \bibinfo{author}{Dean, J.}, \bibinfo{author}{Devin, M.},
  \bibinfo{author}{Ghemawat, S.}, \bibinfo{author}{Goodfellow, I.},
  \bibinfo{author}{Harp, A.}, \bibinfo{author}{Irving, G.},
  \bibinfo{author}{Isard, M.}, \bibinfo{author}{Jia, Y.},
  \bibinfo{author}{Jozefowicz, R.}, \bibinfo{author}{Kaiser, L.},
  \bibinfo{author}{Kudlur, M.}, \bibinfo{author}{Levenberg, J.},
  \bibinfo{author}{Mane, D.}, \bibinfo{author}{Monga, R.},
  \bibinfo{author}{Moore, S.}, \bibinfo{author}{Murray, D.},
  \bibinfo{author}{Olah, C.}, \bibinfo{author}{Schuster, M.},
  \bibinfo{author}{Shlens, J.}, \bibinfo{author}{Steiner, B.},
  \bibinfo{author}{Sutskever, I.}, \bibinfo{author}{Talwar, K.},
  \bibinfo{author}{Tucker, P.}, \bibinfo{author}{Vanhoucke, V.},
  \bibinfo{author}{Vasudevan, V.}, \bibinfo{author}{Viegas, F.},
  \bibinfo{author}{Vinyals, O.}, \bibinfo{author}{Warden, P.},
  \bibinfo{author}{Wattenberg, M.}, \bibinfo{author}{Wicke, M.},
  \bibinfo{author}{Yu, Y.}, \bibinfo{author}{Zheng, X.}, \bibinfo{year}{2016}.
\newblock \bibinfo{title}{{TensorFlow}: {Large-Scale} machine learning on
  heterogeneous distributed systems}
  \href{http://arxiv.org/abs/1603.04467}{{\tt arXiv:1603.04467}}.
%Type = Misc
\bibitem[{{AEMO}(2020)}]{Aemo2020-xx}
\bibinfo{author}{{AEMO}}, \bibinfo{year}{2020}.
\newblock \bibinfo{title}{Data dashboard {NEM}}.
\newblock
  \bibinfo{howpublished}{\url{https://www.aemo.com.au/energy-systems/electricity/national-electricity-market-nem/data-nem/data-dashboard-nem}}.
\newblock \bibinfo{note}{Accessed: 2020-6-30}.
%Type = Article
\bibitem[{Almonacid et~al.(2013)Almonacid, P{\'e}rez-Higueras, Rodrigo and
  Hontoria}]{Almonacid2013-ot}
\bibinfo{author}{Almonacid, F.}, \bibinfo{author}{P{\'e}rez-Higueras, P.},
  \bibinfo{author}{Rodrigo, P.}, \bibinfo{author}{Hontoria, L.},
  \bibinfo{year}{2013}.
\newblock \bibinfo{title}{Generation of ambient temperature hourly time series
  for some spanish locations by artificial neural networks}.
\newblock \bibinfo{journal}{Renew. Energy} \bibinfo{volume}{51},
  \bibinfo{pages}{285--291}.
%Type = Article
\bibitem[{Athanasopoulos et~al.(2018)Athanasopoulos, Song and
  Sun}]{Athanasopoulos2018-bn}
\bibinfo{author}{Athanasopoulos, G.}, \bibinfo{author}{Song, H.},
  \bibinfo{author}{Sun, J.A.}, \bibinfo{year}{2018}.
\newblock \bibinfo{title}{Bagging in tourism demand modeling and forecasting}.
\newblock \bibinfo{journal}{J. Travel Res.} \bibinfo{volume}{57},
  \bibinfo{pages}{52--68}.
%Type = Misc
\bibitem[{{AusGrid}(2019)}]{AusGrid2019-wq}
\bibinfo{author}{{AusGrid}}, \bibinfo{year}{2019}.
\newblock \bibinfo{title}{Innovation and research - ausgrid}.
\newblock
  \bibinfo{howpublished}{\url{https://www.ausgrid.com.au/Industry/Innovation-and-research/}}.
\newblock \bibinfo{note}{Accessed: 2019-5-16}.
%Type = Misc
\bibitem[{{AutoML Group}(2017)}]{AutoML_Group2017-bg}
\bibinfo{author}{{AutoML Group}}, \bibinfo{year}{2017}.
\newblock \bibinfo{title}{Smac v3: Algorithm configuration in python}.
\newblock \bibinfo{howpublished}{\url{https://github.com/automl/SMAC3}}.
\newblock \bibinfo{note}{Accessed: 2020-2-13}.
%Type = Inproceedings
\bibitem[{Bandara et~al.(2020a)Bandara, Bergmeir, Campbell, Scott and
  Lubman}]{Bandara2020-en}
\bibinfo{author}{Bandara, K.}, \bibinfo{author}{Bergmeir, C.},
  \bibinfo{author}{Campbell, S.}, \bibinfo{author}{Scott, D.},
  \bibinfo{author}{Lubman, D.}, \bibinfo{year}{2020}a.
\newblock \bibinfo{title}{Towards accurate predictions and causal 'what-if'
  analyses for planning and policy-making: A case study in emergency medical
  services demand}.
%Type = Article
\bibitem[{Bandara et~al.(2020b)Bandara, Bergmeir and
  Hewamalage}]{Bandara2020-zt}
\bibinfo{author}{Bandara, K.}, \bibinfo{author}{Bergmeir, C.},
  \bibinfo{author}{Hewamalage, H.}, \bibinfo{year}{2020}b.
\newblock \bibinfo{title}{{LSTM-MSNet}: Leveraging forecasts on sets of related
  time series with multiple seasonal patterns}.
\newblock \bibinfo{journal}{IEEE Trans. Neural Netw. Learn. Syst.} ,
  \bibinfo{pages}{1--14}\DOIprefix\doi{10.1109/TNNLS.2020.2985720}.
%Type = Article
\bibitem[{Bandara et~al.(2020c)Bandara, Bergmeir and Smyl}]{Bandara2019-iv}
\bibinfo{author}{Bandara, K.}, \bibinfo{author}{Bergmeir, C.},
  \bibinfo{author}{Smyl, S.}, \bibinfo{year}{2020}c.
\newblock \bibinfo{title}{Forecasting across time series databases using
  recurrent neural networks on groups of similar series: A clustering
  approach}.
\newblock \bibinfo{journal}{Expert Syst. Appl.} \bibinfo{volume}{140},
  \bibinfo{pages}{112896}.
%Type = Inproceedings
\bibitem[{Bandara et~al.(2019)Bandara, Shi, Bergmeir, Hewamalage, Tran and
  Seaman}]{Bandara2019-bg}
\bibinfo{author}{Bandara, K.}, \bibinfo{author}{Shi, P.},
  \bibinfo{author}{Bergmeir, C.}, \bibinfo{author}{Hewamalage, H.},
  \bibinfo{author}{Tran, Q.}, \bibinfo{author}{Seaman, B.},
  \bibinfo{year}{2019}.
\newblock \bibinfo{title}{Sales demand forecast in e-commerce using a long
  {Short-Term} memory neural network methodology}, in:
  \bibinfo{booktitle}{Neural Information Processing},
  \bibinfo{publisher}{Springer International Publishing}. pp.
  \bibinfo{pages}{462--474}.
%Type = Article
\bibitem[{Ben~Taieb et~al.(2012)Ben~Taieb, Bontempi, Atiya and
  Sorjamaa}]{Ben_Taieb2012-re}
\bibinfo{author}{Ben~Taieb, S.}, \bibinfo{author}{Bontempi, G.},
  \bibinfo{author}{Atiya, A.F.}, \bibinfo{author}{Sorjamaa, A.},
  \bibinfo{year}{2012}.
\newblock \bibinfo{title}{A review and comparison of strategies for multi-step
  ahead time series forecasting based on the {NN5} forecasting competition}.
\newblock \bibinfo{journal}{Expert Syst. Appl.} \bibinfo{volume}{39},
  \bibinfo{pages}{7067--7083}.
%Type = Inproceedings
\bibitem[{Bengio(2012)}]{Bengio2012-tr}
\bibinfo{author}{Bengio, Y.}, \bibinfo{year}{2012}.
\newblock \bibinfo{title}{Deep learning of representations for unsupervised and
  transfer learning}, in: \bibinfo{editor}{Guyon, I.}, \bibinfo{editor}{Dror,
  G.}, \bibinfo{editor}{Lemaire, V.}, \bibinfo{editor}{Taylor, G.},
  \bibinfo{editor}{Silver, D.} (Eds.), \bibinfo{booktitle}{Proceedings of
  {ICML} Workshop on Unsupervised and Transfer Learning},
  \bibinfo{publisher}{PMLR}, \bibinfo{address}{Bellevue, Washington, USA}. pp.
  \bibinfo{pages}{17--36}.
%Type = Article
\bibitem[{Bengio et~al.(1994)Bengio, Simard and Frasconi}]{Bengio1994-oq}
\bibinfo{author}{Bengio, Y.}, \bibinfo{author}{Simard, P.},
  \bibinfo{author}{Frasconi, P.}, \bibinfo{year}{1994}.
\newblock \bibinfo{title}{Learning long-term dependencies with gradient descent
  is difficult}.
\newblock \bibinfo{journal}{IEEE Trans. Neural Netw.} \bibinfo{volume}{5},
  \bibinfo{pages}{157--166}.
%Type = Article
\bibitem[{Bergmeir et~al.(2016)Bergmeir, Hyndman and
  Ben{\'\i}tez}]{Bergmeir2016-zk}
\bibinfo{author}{Bergmeir, C.}, \bibinfo{author}{Hyndman, R.J.},
  \bibinfo{author}{Ben{\'\i}tez, J.M.}, \bibinfo{year}{2016}.
\newblock \bibinfo{title}{Bagging exponential smoothing methods using {STL}
  decomposition and {Box--Cox} transformation}.
\newblock \bibinfo{journal}{Int. J. Forecast.} \bibinfo{volume}{32},
  \bibinfo{pages}{303--312}.
%Type = Article
\bibitem[{Borovykh et~al.(2017)Borovykh, Bohte and Oosterlee}]{Borovykh2017-vz}
\bibinfo{author}{Borovykh, A.}, \bibinfo{author}{Bohte, S.},
  \bibinfo{author}{Oosterlee, C.W.}, \bibinfo{year}{2017}.
\newblock \bibinfo{title}{Conditional time series forecasting with
  convolutional neural networks} \href{http://arxiv.org/abs/1703.04691}{{\tt
  arXiv:1703.04691}}.
%Type = Book
\bibitem[{Box et~al.(2015)Box, Jenkins, Reinsel and Ljung}]{Box2015-bz}
\bibinfo{author}{Box, G.E.P.}, \bibinfo{author}{Jenkins, G.M.},
  \bibinfo{author}{Reinsel, G.C.}, \bibinfo{author}{Ljung, G.M.},
  \bibinfo{year}{2015}.
\newblock \bibinfo{title}{Time Series Analysis: Forecasting and Control}.
\newblock \bibinfo{publisher}{John Wiley \& Sons}.
%Type = Article
\bibitem[{Chen et~al.(2020)Chen, Kang, Chen and Wang}]{kang2020tcn}
\bibinfo{author}{Chen, Y.}, \bibinfo{author}{Kang, Y.}, \bibinfo{author}{Chen,
  Y.}, \bibinfo{author}{Wang, Z.}, \bibinfo{year}{2020}.
\newblock \bibinfo{title}{Probabilistic forecasting with temporal convolutional
  neural network}.
\newblock \bibinfo{journal}{Neurocomputing} \bibinfo{volume}{399},
  \bibinfo{pages}{491--501}.
%Type = Inproceedings
\bibitem[{Chen et~al.(2018)Chen, Li and Zhang}]{Chen2018-wo}
\bibinfo{author}{Chen, Y.}, \bibinfo{author}{Li, P.}, \bibinfo{author}{Zhang,
  B.}, \bibinfo{year}{2018}.
\newblock \bibinfo{title}{Bayesian renewables scenario generation via deep
  generative networks}, in: \bibinfo{booktitle}{2018 52nd Annual Conference on
  Information Sciences and Systems ({CISS})}, pp. \bibinfo{pages}{1--6}.
%Type = Article
\bibitem[{Cho et~al.(2014)Cho, van Merrienboer, Gulcehre, Bahdanau, Bougares,
  Schwenk and Bengio}]{Cho2014-vr}
\bibinfo{author}{Cho, K.}, \bibinfo{author}{van Merrienboer, B.},
  \bibinfo{author}{Gulcehre, C.}, \bibinfo{author}{Bahdanau, D.},
  \bibinfo{author}{Bougares, F.}, \bibinfo{author}{Schwenk, H.},
  \bibinfo{author}{Bengio, Y.}, \bibinfo{year}{2014}.
\newblock \bibinfo{title}{Learning phrase representations using {RNN}
  {Encoder-Decoder} for statistical machine translation}
  \href{http://arxiv.org/abs/1406.1078}{{\tt arXiv:1406.1078}}.
%Type = Article
\bibitem[{Cleveland et~al.(1990)Cleveland, Cleveland and
  Terpenning}]{Cleveland1990-rc}
\bibinfo{author}{Cleveland, R.B.}, \bibinfo{author}{Cleveland, W.S.},
  \bibinfo{author}{Terpenning, I.}, \bibinfo{year}{1990}.
\newblock \bibinfo{title}{{STL}: A seasonal-trend decomposition procedure based
  on loess}.
\newblock \bibinfo{journal}{J. Off. Stat.} \bibinfo{volume}{6},
  \bibinfo{pages}{3--73}.
%Type = Misc
\bibitem[{Crone(2008)}]{Crone2008-ye}
\bibinfo{author}{Crone, S.F.}, \bibinfo{year}{2008}.
\newblock \bibinfo{title}{{NN5} competition}.
\newblock
  \bibinfo{howpublished}{\url{http://www.neural-forecasting-competition.com/NN5/}}.
\newblock \bibinfo{note}{Accessed: 2017-8-18}.
%Type = Article
\bibitem[{Crone et~al.(2011)Crone, Hibon and Nikolopoulos}]{Crone2011-vv}
\bibinfo{author}{Crone, S.F.}, \bibinfo{author}{Hibon, M.},
  \bibinfo{author}{Nikolopoulos, K.}, \bibinfo{year}{2011}.
\newblock \bibinfo{title}{Advances in forecasting with neural networks?
  empirical evidence from the {NN3} competition on time series prediction}.
\newblock \bibinfo{journal}{Int. J. Forecast.} \bibinfo{volume}{27},
  \bibinfo{pages}{635--660}.
%Type = Article
\bibitem[{Dantas et~al.(2017)Dantas, Cyrino~Oliveira and
  Varela~Repolho}]{Dantas2017-rv}
\bibinfo{author}{Dantas, T.M.}, \bibinfo{author}{Cyrino~Oliveira, F.L.},
  \bibinfo{author}{Varela~Repolho, H.M.}, \bibinfo{year}{2017}.
\newblock \bibinfo{title}{Air transportation demand forecast through bagging
  holt winters methods}.
\newblock \bibinfo{journal}{J. Air Transp. Manage.} \bibinfo{volume}{59},
  \bibinfo{pages}{116--123}.
%Type = Inproceedings
\bibitem[{Denaxas et~al.(2015)Denaxas, Bandyopadhyay, Pati{\~n}o-Echeverri and
  Pitsianis}]{Denaxas2015-sl}
\bibinfo{author}{Denaxas, E.A.}, \bibinfo{author}{Bandyopadhyay, R.},
  \bibinfo{author}{Pati{\~n}o-Echeverri, D.}, \bibinfo{author}{Pitsianis, N.},
  \bibinfo{year}{2015}.
\newblock \bibinfo{title}{{SynTiSe}: A modified multi-regime {MCMC} approach
  for generation of wind power synthetic time series}, in:
  \bibinfo{booktitle}{2015 Annual {IEEE} Systems Conference ({SysCon})
  Proceedings}, pp. \bibinfo{pages}{668--674}.
%Type = Article
\bibitem[{Donyavi and Asadi(2020)}]{DONYAVI2020107543}
\bibinfo{author}{Donyavi, Z.}, \bibinfo{author}{Asadi, S.},
  \bibinfo{year}{2020}.
\newblock \bibinfo{title}{Diverse training dataset generation based on a
  multi-objective optimization for semi-supervised classification}.
\newblock \bibinfo{journal}{Pattern Recognit.} \bibinfo{volume}{108},
  \bibinfo{pages}{107543}.
%Type = Article
\bibitem[{Elman(1990)}]{Elman1990-my}
\bibinfo{author}{Elman, J.L.}, \bibinfo{year}{1990}.
\newblock \bibinfo{title}{Finding structure in time}.
\newblock \bibinfo{journal}{Cogn. Sci.} \bibinfo{volume}{14},
  \bibinfo{pages}{179--211}.
%Type = Article
\bibitem[{Esteban et~al.(2017)Esteban, Hyland and R{\"a}tsch}]{Esteban2017-yg}
\bibinfo{author}{Esteban, C.}, \bibinfo{author}{Hyland, S.L.},
  \bibinfo{author}{R{\"a}tsch, G.}, \bibinfo{year}{2017}.
\newblock \bibinfo{title}{Real-valued (medical) time series generation with
  recurrent conditional {GANs}} \href{http://arxiv.org/abs/1706.02633}{{\tt
  arXiv:1706.02633}}.
%Type = Article
\bibitem[{Fawaz et~al.(2018)Fawaz, Forestier, Weber, Idoumghar and
  Muller}]{Fawaz2018-fj}
\bibinfo{author}{Fawaz, H.I.}, \bibinfo{author}{Forestier, G.},
  \bibinfo{author}{Weber, J.}, \bibinfo{author}{Idoumghar, L.},
  \bibinfo{author}{Muller, P.A.}, \bibinfo{year}{2018}.
\newblock \bibinfo{title}{Data augmentation using synthetic data for time
  series classification with deep residual networks}
  \href{http://arxiv.org/abs/1808.02455}{{\tt arXiv:1808.02455}}.
%Type = Inproceedings
\bibitem[{Forestier et~al.(2017)Forestier, Petitjean, Dau, Webb and
  Keogh}]{Forestier2017-su}
\bibinfo{author}{Forestier, G.}, \bibinfo{author}{Petitjean, F.},
  \bibinfo{author}{Dau, H.A.}, \bibinfo{author}{Webb, G.I.},
  \bibinfo{author}{Keogh, E.}, \bibinfo{year}{2017}.
\newblock \bibinfo{title}{Generating synthetic time series to augment sparse
  datasets}, in: \bibinfo{booktitle}{2017 {IEEE} International Conference on
  Data Mining ({ICDM})}, pp. \bibinfo{pages}{865--870}.
%Type = Article
\bibitem[{Fu et~al.(2019)Fu, Chen, Zeng, Zhuang and Sudjianto}]{Fu2019-mw}
\bibinfo{author}{Fu, R.}, \bibinfo{author}{Chen, J.}, \bibinfo{author}{Zeng,
  S.}, \bibinfo{author}{Zhuang, Y.}, \bibinfo{author}{Sudjianto, A.},
  \bibinfo{year}{2019}.
\newblock \bibinfo{title}{Time series simulation by conditional generative
  adversarial net} \href{http://arxiv.org/abs/1904.11419}{{\tt
  arXiv:1904.11419}}.
%Type = Article
\bibitem[{Garc{\'\i}a et~al.(2010)Garc{\'\i}a, Fern{\'a}ndez, Luengo and
  Herrera}]{Garcia2010-jx}
\bibinfo{author}{Garc{\'\i}a, S.}, \bibinfo{author}{Fern{\'a}ndez, A.},
  \bibinfo{author}{Luengo, J.}, \bibinfo{author}{Herrera, F.},
  \bibinfo{year}{2010}.
\newblock \bibinfo{title}{Advanced nonparametric tests for multiple comparisons
  in the design of experiments in computational intelligence and data mining:
  Experimental analysis of power}.
\newblock \bibinfo{journal}{Inf. Sci.} \bibinfo{volume}{180},
  \bibinfo{pages}{2044--2064}.
%Type = Inproceedings
\bibitem[{Glorot et~al.(2011)Glorot, Bordes and Bengio}]{Glorot2011-kp}
\bibinfo{author}{Glorot, X.}, \bibinfo{author}{Bordes, A.},
  \bibinfo{author}{Bengio, Y.}, \bibinfo{year}{2011}.
\newblock \bibinfo{title}{Domain adaptation for large-scale sentiment
  classification: A deep learning approach}, in:
  \bibinfo{booktitle}{Proceedings of the 28th International Conference on
  International Conference on Machine Learning},
  \bibinfo{publisher}{Omnipress}, \bibinfo{address}{USA}. pp.
  \bibinfo{pages}{513--520}.
%Type = Incollection
\bibitem[{Goodfellow et~al.(2014)Goodfellow, Pouget-Abadie, Mirza, Xu,
  Warde-Farley, Ozair, Courville and Bengio}]{Goodfellow2014-kb}
\bibinfo{author}{Goodfellow, I.}, \bibinfo{author}{Pouget-Abadie, J.},
  \bibinfo{author}{Mirza, M.}, \bibinfo{author}{Xu, B.},
  \bibinfo{author}{Warde-Farley, D.}, \bibinfo{author}{Ozair, S.},
  \bibinfo{author}{Courville, A.}, \bibinfo{author}{Bengio, Y.},
  \bibinfo{year}{2014}.
\newblock \bibinfo{title}{Generative adversarial nets}, in:
  \bibinfo{editor}{Ghahramani, Z.}, \bibinfo{editor}{Welling, M.},
  \bibinfo{editor}{Cortes, C.}, \bibinfo{editor}{Lawrence, N.D.},
  \bibinfo{editor}{Weinberger, K.Q.} (Eds.), \bibinfo{booktitle}{Advances in
  Neural Information Processing Systems 27}. \bibinfo{publisher}{Curran
  Associates, Inc.}, pp. \bibinfo{pages}{2672--2680}.
%Type = Article
\bibitem[{Gupta et~al.(2018)Gupta, Malhotra, Vig and Shroff}]{Gupta2018-no}
\bibinfo{author}{Gupta, P.}, \bibinfo{author}{Malhotra, P.},
  \bibinfo{author}{Vig, L.}, \bibinfo{author}{Shroff, G.},
  \bibinfo{year}{2018}.
\newblock \bibinfo{title}{Transfer learning for clinical time series analysis
  using recurrent neural networks} \href{http://arxiv.org/abs/1807.01705}{{\tt
  arXiv:1807.01705}}.
%Type = Article
\bibitem[{Hannun et~al.(2014)Hannun, Case, Casper, Catanzaro, Diamos, Elsen,
  Prenger, Satheesh, Sengupta, Coates and Ng}]{Hannun2014-ju}
\bibinfo{author}{Hannun, A.}, \bibinfo{author}{Case, C.},
  \bibinfo{author}{Casper, J.}, \bibinfo{author}{Catanzaro, B.},
  \bibinfo{author}{Diamos, G.}, \bibinfo{author}{Elsen, E.},
  \bibinfo{author}{Prenger, R.}, \bibinfo{author}{Satheesh, S.},
  \bibinfo{author}{Sengupta, S.}, \bibinfo{author}{Coates, A.},
  \bibinfo{author}{Ng, A.Y.}, \bibinfo{year}{2014}.
\newblock \bibinfo{title}{Deep speech: Scaling up end-to-end speech
  recognition} \href{http://arxiv.org/abs/1412.5567}{{\tt arXiv:1412.5567}}.
%Type = Inproceedings
\bibitem[{He et~al.(2016)He, Zhang, Ren and Sun}]{He2016-wm}
\bibinfo{author}{He, K.}, \bibinfo{author}{Zhang, X.}, \bibinfo{author}{Ren,
  S.}, \bibinfo{author}{Sun, J.}, \bibinfo{year}{2016}.
\newblock \bibinfo{title}{Deep residual learning for image recognition}, in:
  \bibinfo{booktitle}{2016 {IEEE} Conference on Computer Vision and Pattern
  Recognition ({CVPR})}, pp. \bibinfo{pages}{770--778}.
%Type = Article
\bibitem[{Hewamalage et~al.(2020)Hewamalage, Bergmeir and
  Bandara}]{Hewamalage2019-il}
\bibinfo{author}{Hewamalage, H.}, \bibinfo{author}{Bergmeir, C.},
  \bibinfo{author}{Bandara, K.}, \bibinfo{year}{2020}.
\newblock \bibinfo{title}{Recurrent neural networks for time series
  forecasting: Current status and future directions}.
\newblock \bibinfo{journal}{International Journal of Forecasting (in press)} .
%Type = Article
\bibitem[{Huang et~al.(2020)Huang, Xu, Li, Shen and Chen}]{Huang2020-fv}
\bibinfo{author}{Huang, B.}, \bibinfo{author}{Xu, T.}, \bibinfo{author}{Li,
  J.}, \bibinfo{author}{Shen, Z.}, \bibinfo{author}{Chen, Y.},
  \bibinfo{year}{2020}.
\newblock \bibinfo{title}{Transfer learning-based discriminative correlation
  filter for visual tracking}.
\newblock \bibinfo{journal}{Pattern Recognit.} \bibinfo{volume}{100},
  \bibinfo{pages}{107157}.
%Type = Inproceedings
\bibitem[{Hutter et~al.(2011)Hutter, Hoos and Leyton-Brown}]{Hutter2011-wa}
\bibinfo{author}{Hutter, F.}, \bibinfo{author}{Hoos, H.H.},
  \bibinfo{author}{Leyton-Brown, K.}, \bibinfo{year}{2011}.
\newblock \bibinfo{title}{Sequential model-based optimization for general
  algorithm configuration}, in: \bibinfo{booktitle}{Proceedings of the 5th
  International Conference on Learning and Intelligent Optimization},
  \bibinfo{publisher}{Springer-Verlag}, \bibinfo{address}{Berlin, Heidelberg}.
  pp. \bibinfo{pages}{507--523}.
%Type = Manual
\bibitem[{Hyndman et~al.(2019)Hyndman, Athanasopoulos, Bergmeir, Caceres,
  Chhay, O'Hara-Wild, Petropoulos, Razbash, Wang and Yasmeen}]{Hyndman2015-vm}
\bibinfo{author}{Hyndman, R.J.}, \bibinfo{author}{Athanasopoulos, G.},
  \bibinfo{author}{Bergmeir, C.}, \bibinfo{author}{Caceres, G.},
  \bibinfo{author}{Chhay, L.}, \bibinfo{author}{O'Hara-Wild, M.},
  \bibinfo{author}{Petropoulos, F.}, \bibinfo{author}{Razbash, S.},
  \bibinfo{author}{Wang, E.}, \bibinfo{author}{Yasmeen, F.},
  \bibinfo{year}{2019}.
\newblock \bibinfo{title}{{forecast}: Forecasting functions for time series and
  linear models}.
\newblock \URLprefix \url{http://pkg.robjhyndman.com/forecast}.
  \bibinfo{note}{{R} package version 8.5}.
%Type = Article
\bibitem[{Hyndman and Khandakar(2008)}]{Khandakar2008-hd}
\bibinfo{author}{Hyndman, R.J.}, \bibinfo{author}{Khandakar, Y.},
  \bibinfo{year}{2008}.
\newblock \bibinfo{title}{Automatic time series forecasting: the forecast
  package for {R}}.
\newblock \bibinfo{journal}{J. Stat. Softw.} \bibinfo{volume}{27}.
%Type = Article
\bibitem[{Hyndman and Koehler(2006)}]{Hyndman2006-ue}
\bibinfo{author}{Hyndman, R.J.}, \bibinfo{author}{Koehler, A.B.},
  \bibinfo{year}{2006}.
\newblock \bibinfo{title}{Another look at measures of forecast accuracy}.
\newblock \bibinfo{journal}{Int. J. Forecast.} .
%Type = Book
\bibitem[{Hyndman et~al.(2008)Hyndman, Koehler, Keith~Ord and
  Snyder}]{Hyndman2008-yd}
\bibinfo{author}{Hyndman, R.J.}, \bibinfo{author}{Koehler, A.B.},
  \bibinfo{author}{Keith~Ord, J.}, \bibinfo{author}{Snyder, R.D.},
  \bibinfo{year}{2008}.
\newblock \bibinfo{title}{Forecasting with Exponential Smoothing: The State
  Space Approach}.
\newblock \bibinfo{publisher}{Springer Science \& Business Media}.
%Type = Inproceedings
\bibitem[{Iftikhar et~al.(2017)Iftikhar, Liu, Danalachi, Nordbjerg and
  Vollesen}]{Iftikhar2017-wy}
\bibinfo{author}{Iftikhar, N.}, \bibinfo{author}{Liu, X.},
  \bibinfo{author}{Danalachi, S.}, \bibinfo{author}{Nordbjerg, F.E.},
  \bibinfo{author}{Vollesen, J.H.}, \bibinfo{year}{2017}.
\newblock \bibinfo{title}{A scalable smart meter data generator using spark},
  in: \bibinfo{booktitle}{On the Move to Meaningful Internet Systems. {OTM}
  2017 Conferences}, \bibinfo{publisher}{Springer International Publishing}.
  pp. \bibinfo{pages}{21--36}.
%Type = Article
\bibitem[{Januschowski et~al.(2020)Januschowski, Gasthaus, Wang, Salinas,
  Flunkert, Bohlke-Schneider and Callot}]{Januschowski2020-ud}
\bibinfo{author}{Januschowski, T.}, \bibinfo{author}{Gasthaus, J.},
  \bibinfo{author}{Wang, Y.}, \bibinfo{author}{Salinas, D.},
  \bibinfo{author}{Flunkert, V.}, \bibinfo{author}{Bohlke-Schneider, M.},
  \bibinfo{author}{Callot, L.}, \bibinfo{year}{2020}.
\newblock \bibinfo{title}{Criteria for classifying forecasting methods}.
\newblock \bibinfo{journal}{Int. J. Forecast.} \bibinfo{volume}{36},
  \bibinfo{pages}{167--177}.
%Type = Article
\bibitem[{Kang et~al.(2020a)Kang, Hyndman and Li}]{Kang2019-dy}
\bibinfo{author}{Kang, Y.}, \bibinfo{author}{Hyndman, R.J.},
  \bibinfo{author}{Li, F.}, \bibinfo{year}{2020}a.
\newblock \bibinfo{title}{{GRATIS}: {GeneRAting} {TIme} series with diverse and
  controllable characteristics}.
\newblock \bibinfo{journal}{Stat. Anal. Data Min.} \bibinfo{volume}{13},
  \bibinfo{pages}{354--376}.
%Type = Manual
\bibitem[{Kang et~al.(2020b)Kang, O'Hara-Wild, Hyndman and Li}]{Kang2018-wm}
\bibinfo{author}{Kang, Y.}, \bibinfo{author}{O'Hara-Wild, M.},
  \bibinfo{author}{Hyndman, R.J.}, \bibinfo{author}{Li, F.},
  \bibinfo{year}{2020}b.
\newblock \bibinfo{title}{{GRATIS}: {GeneRAting} {TIme} Series with diverse and
  controllable characteristics}.
\newblock \URLprefix \url{https://github.com/ykang/gratis}.
  \bibinfo{note}{accessed: 2020-2-11}.
%Type = Inproceedings
\bibitem[{Kegel et~al.(2018)Kegel, Hahmann and Lehner}]{Kegel2018-rs}
\bibinfo{author}{Kegel, L.}, \bibinfo{author}{Hahmann, M.},
  \bibinfo{author}{Lehner, W.}, \bibinfo{year}{2018}.
\newblock \bibinfo{title}{Feature-based comparison and generation of time
  series}, in: \bibinfo{booktitle}{Proceedings of the 30th International
  Conference on Scientific and Statistical Database Management - {SSDBM} '18},
  \bibinfo{publisher}{ACM Press}, \bibinfo{address}{New York, New York, USA}.
  pp. \bibinfo{pages}{1--12}.
%Type = Incollection
\bibitem[{Krizhevsky et~al.(2012)Krizhevsky, Sutskever and
  Hinton}]{Krizhevsky2012-ct}
\bibinfo{author}{Krizhevsky, A.}, \bibinfo{author}{Sutskever, I.},
  \bibinfo{author}{Hinton, G.E.}, \bibinfo{year}{2012}.
\newblock \bibinfo{title}{{ImageNet} classification with deep convolutional
  neural networks}, in: \bibinfo{editor}{Pereira, F.}, \bibinfo{editor}{Burges,
  C.J.C.}, \bibinfo{editor}{Bottou, L.}, \bibinfo{editor}{Weinberger, K.Q.}
  (Eds.), \bibinfo{booktitle}{Advances in Neural Information Processing Systems
  25}. \bibinfo{publisher}{Curran Associates, Inc.}, pp.
  \bibinfo{pages}{1097--1105}.
%Type = Misc
\bibitem[{Lai(2018)}]{Lai2018-yb}
\bibinfo{author}{Lai, G.}, \bibinfo{year}{2018}.
\newblock \bibinfo{title}{Multivariate time series forecasting}.
\newblock
  \bibinfo{howpublished}{\url{https://github.com/laiguokun/multivariate-time-series-data}}.
\newblock \bibinfo{note}{Accessed: 2020-6-30}.
%Type = Inproceedings
\bibitem[{Lai et~al.(2018)Lai, Chang, Yang and Liu}]{Lai2018-zx}
\bibinfo{author}{Lai, G.}, \bibinfo{author}{Chang, W.C.},
  \bibinfo{author}{Yang, Y.}, \bibinfo{author}{Liu, H.}, \bibinfo{year}{2018}.
\newblock \bibinfo{title}{Modeling long- and {Short-Term} temporal patterns
  with deep neural networks}, in: \bibinfo{booktitle}{The 41st International
  {ACM} {SIGIR} Conference on Research \& Development in Information
  Retrieval}, \bibinfo{publisher}{ACM}, \bibinfo{address}{New York, NY, USA}.
  pp. \bibinfo{pages}{95--104}.
%Type = Inproceedings
\bibitem[{Laptev et~al.(2018)Laptev, Yu and Rajagopal}]{Laptev2018-kb}
\bibinfo{author}{Laptev, N.}, \bibinfo{author}{Yu, J.},
  \bibinfo{author}{Rajagopal, R.}, \bibinfo{year}{2018}.
\newblock \bibinfo{title}{Reconstruction and regression loss for time-series
  transfer learning}, in: \bibinfo{booktitle}{Proceedings of the Special
  Interest Group on Knowledge Discovery and Data Mining (SIGKDD) and the 4th
  Workshop on the Mining and LEarning from Time Series (MiLeTS), London, UK}.
%Type = Inproceedings
\bibitem[{Le~Guennec et~al.(2016)Le~Guennec, Malinowski and
  Tavenard}]{Le_Guennec2016-tc}
\bibinfo{author}{Le~Guennec, A.}, \bibinfo{author}{Malinowski, S.},
  \bibinfo{author}{Tavenard, R.}, \bibinfo{year}{2016}.
\newblock \bibinfo{title}{Data augmentation for time series classification
  using convolutional neural networks}, in: \bibinfo{booktitle}{{ECML/PKDD}
  workshop on advanced analytics and learning on temporal data}.
%Type = Article
\bibitem[{Li et~al.(2020a)Li, Grandvalet and Davoine}]{Li2020-og}
\bibinfo{author}{Li, X.}, \bibinfo{author}{Grandvalet, Y.},
  \bibinfo{author}{Davoine, F.}, \bibinfo{year}{2020}a.
\newblock \bibinfo{title}{A baseline regularization scheme for transfer
  learning with convolutional neural networks}.
\newblock \bibinfo{journal}{Pattern Recognit.} \bibinfo{volume}{98},
  \bibinfo{pages}{107049}.
%Type = Article
\bibitem[{Li et~al.(2020b)Li, Kang and Li}]{li2020imaging}
\bibinfo{author}{Li, X.}, \bibinfo{author}{Kang, Y.}, \bibinfo{author}{Li, F.},
  \bibinfo{year}{2020}b.
\newblock \bibinfo{title}{Forecasting with time series imaging}.
\newblock \bibinfo{journal}{Expert Syst. Appl.} \bibinfo{volume}{160},
  \bibinfo{pages}{113680}.
%Type = Article
\bibitem[{Makridakis et~al.(2018)Makridakis, Spiliotis and
  Assimakopoulos}]{Makridakis2018-cm}
\bibinfo{author}{Makridakis, S.}, \bibinfo{author}{Spiliotis, E.},
  \bibinfo{author}{Assimakopoulos, V.}, \bibinfo{year}{2018}.
\newblock \bibinfo{title}{The {M4} competition: Results, findings, conclusion
  and way forward}.
\newblock \bibinfo{journal}{Int. J. Forecast.} \bibinfo{volume}{34},
  \bibinfo{pages}{802--808}.
%Type = Inproceedings
\bibitem[{Mikolov et~al.(2010)Mikolov, Karafi{\'a}t, Burget, Cernocky and
  Khudanpur}]{Mikolov2010-rb}
\bibinfo{author}{Mikolov, T.}, \bibinfo{author}{Karafi{\'a}t, M.},
  \bibinfo{author}{Burget, L.}, \bibinfo{author}{Cernocky, J.},
  \bibinfo{author}{Khudanpur, S.}, \bibinfo{year}{2010}.
\newblock \bibinfo{title}{Recurrent neural network based language model}, in:
  \bibinfo{booktitle}{Interspeech}, \bibinfo{publisher}{fit.vutbr.cz}.
  p.~\bibinfo{pages}{3}.
%Type = Misc
\bibitem[{Orabona(2017)}]{Orabona2017-lj}
\bibinfo{author}{Orabona, F.}, \bibinfo{year}{2017}.
\newblock \bibinfo{title}{cocob}.
\newblock \bibinfo{howpublished}{\url{https://github.com/bremen79/cocob}}.
\newblock \bibinfo{note}{Accessed: 2020-2-13}.
%Type = Inproceedings
\bibitem[{Orabona and Tommasi(2017)}]{Orabona2017-ij}
\bibinfo{author}{Orabona, F.}, \bibinfo{author}{Tommasi, T.},
  \bibinfo{year}{2017}.
\newblock \bibinfo{title}{Training deep networks without learning rates through
  coin betting}, in: \bibinfo{booktitle}{Proceedings of the 31st International
  Conference on Neural Information Processing Systems},
  \bibinfo{publisher}{Curran Associates Inc.}, \bibinfo{address}{USA}. pp.
  \bibinfo{pages}{2157--2167}.
%Type = Article
\bibitem[{Pan and Yang(2010)}]{Pan2010-fi}
\bibinfo{author}{Pan, S.J.}, \bibinfo{author}{Yang, Q.}, \bibinfo{year}{2010}.
\newblock \bibinfo{title}{A survey on transfer learning}.
\newblock \bibinfo{journal}{IEEE Trans. Knowl. Data Eng.} \bibinfo{volume}{22},
  \bibinfo{pages}{1345--1359}.
%Type = Article
\bibitem[{Papaefthymiou and Klockl(2008)}]{Papaefthymiou2008-va}
\bibinfo{author}{Papaefthymiou, G.}, \bibinfo{author}{Klockl, B.},
  \bibinfo{year}{2008}.
\newblock \bibinfo{title}{{MCMC} for wind power simulation}.
\newblock \bibinfo{journal}{IEEE Trans. Energy Convers.} \bibinfo{volume}{23},
  \bibinfo{pages}{234--240}.
%Type = Misc
\bibitem[{Petitjean(2017)}]{Petitjean2017-gp}
\bibinfo{author}{Petitjean, F.}, \bibinfo{year}{2017}.
\newblock \bibinfo{title}{{DBA}: Averaging for dynamic time warping}.
\newblock \bibinfo{howpublished}{\url{https://github.com/fpetitjean/DBA}}.
\newblock \bibinfo{note}{Accessed: 2020-6-17}.
%Type = Inproceedings
\bibitem[{Purushotham et~al.(2017)Purushotham, Carvalho, Nilanon and
  Liu}]{Purushotham2017-sz}
\bibinfo{author}{Purushotham, S.}, \bibinfo{author}{Carvalho, W.},
  \bibinfo{author}{Nilanon, T.}, \bibinfo{author}{Liu, Y.},
  \bibinfo{year}{2017}.
\newblock \bibinfo{title}{Variational recurrent adversarial deep domain
  adaptation}, in: \bibinfo{booktitle}{{ICLR}}.
%Type = Manual
\bibitem[{{R Core Team}(2013)}]{R_Core_Team2013-bo}
\bibinfo{author}{{R Core Team}}, \bibinfo{year}{2013}.
\newblock \bibinfo{title}{R: A Language and Environment for Statistical
  Computing}.
\newblock \bibinfo{organization}{R Foundation for Statistical Computing}.
  \bibinfo{address}{Vienna, Austria}.
%Type = Inproceedings
\bibitem[{Ramachandran et~al.(2017)Ramachandran, Liu and
  Le}]{Ramachandran2017-px}
\bibinfo{author}{Ramachandran, P.}, \bibinfo{author}{Liu, P.J.},
  \bibinfo{author}{Le, Q.}, \bibinfo{year}{2017}.
\newblock \bibinfo{title}{Unsupervised pretraining for sequence to sequence
  learning}, in: \bibinfo{booktitle}{Proceedings of the 2017 Conference on
  Empirical Methods in Natural Language Processing}, pp.
  \bibinfo{pages}{383--391}.
%Type = Article
\bibitem[{Ribeiro et~al.(2018)Ribeiro, Grolinger, ElYamany, Higashino and
  Capretz}]{Ribeiro2018-aa}
\bibinfo{author}{Ribeiro, M.}, \bibinfo{author}{Grolinger, K.},
  \bibinfo{author}{ElYamany, H.F.}, \bibinfo{author}{Higashino, W.A.},
  \bibinfo{author}{Capretz, M.A.M.}, \bibinfo{year}{2018}.
\newblock \bibinfo{title}{Transfer learning with seasonal and trend adjustment
  for cross-building energy forecasting}.
\newblock \bibinfo{journal}{Energy Build.} \bibinfo{volume}{165},
  \bibinfo{pages}{352--363}.
%Type = Article
\bibitem[{Salinas et~al.(2020)Salinas, Flunkert, Gasthaus and
  Januschowski}]{Salinas2019-dl}
\bibinfo{author}{Salinas, D.}, \bibinfo{author}{Flunkert, V.},
  \bibinfo{author}{Gasthaus, J.}, \bibinfo{author}{Januschowski, T.},
  \bibinfo{year}{2020}.
\newblock \bibinfo{title}{{DeepAR}: Probabilistic forecasting with
  autoregressive recurrent networks}.
\newblock \bibinfo{journal}{International Journal of Forecasting}
  \bibinfo{volume}{36}, \bibinfo{pages}{1181--1191}.
%Type = Article
\bibitem[{Smyl(2020)}]{Smyl2019-cb}
\bibinfo{author}{Smyl, S.}, \bibinfo{year}{2020}.
\newblock \bibinfo{title}{A hybrid method of exponential smoothing and
  recurrent neural networks for time series forecasting}.
\newblock \bibinfo{journal}{International Journal of Forecasting}
  \bibinfo{volume}{36}, \bibinfo{pages}{75--85}.
%Type = Inproceedings
\bibitem[{Smyl and Kuber(2016)}]{Smyl2016-ux}
\bibinfo{author}{Smyl, S.}, \bibinfo{author}{Kuber, K.}, \bibinfo{year}{2016}.
\newblock \bibinfo{title}{Data preprocessing and augmentation for multiple
  short time series forecasting with recurrent neural networks}, in:
  \bibinfo{booktitle}{36th International Symposium on Forecasting}.
%Type = Inproceedings
\bibitem[{{\v S}t{\v e}pni{\v c}ka and Burda(2016)}]{Stepnicka2016-bu}
\bibinfo{author}{{\v S}t{\v e}pni{\v c}ka, M.}, \bibinfo{author}{Burda, M.},
  \bibinfo{year}{2016}.
\newblock \bibinfo{title}{Computational intelligence in forecasting ({CIF})
  2016 time series forecasting competition}, in: \bibinfo{booktitle}{{IEEE}
  {WCCI} 2016, {JCNN-13} Advances in Computational Intelligence for Applied
  Time Series Forecasting ({ACIATSF})}.
%Type = Misc
\bibitem[{Suilin(2018)}]{Suilin2018-tc}
\bibinfo{author}{Suilin, A.}, \bibinfo{year}{2018}.
\newblock \bibinfo{title}{Kaggle-web-traffic}.
\newblock
  \bibinfo{howpublished}{\url{https://github.com/Arturus/kaggle-web-traffic}}.
\newblock \bibinfo{note}{Accessed: 2020-2-10}.
%Type = Incollection
\bibitem[{Sutskever et~al.(2014)Sutskever, Vinyals and Le}]{Sutskever2014-vp}
\bibinfo{author}{Sutskever, I.}, \bibinfo{author}{Vinyals, O.},
  \bibinfo{author}{Le, Q.V.}, \bibinfo{year}{2014}.
\newblock \bibinfo{title}{Sequence to sequence learning with neural networks},
  in: \bibinfo{editor}{Ghahramani, Z.}, \bibinfo{editor}{Welling, M.},
  \bibinfo{editor}{Cortes, C.}, \bibinfo{editor}{Lawrence, N.D.},
  \bibinfo{editor}{Weinberger, K.Q.} (Eds.), \bibinfo{booktitle}{Advances in
  Neural Information Processing Systems 27}. \bibinfo{publisher}{Curran
  Associates, Inc.}, pp. \bibinfo{pages}{3104--3112}.
%Type = Manual
\bibitem[{Svetunkov(2020)}]{Svetunkov2017-je}
\bibinfo{author}{Svetunkov, I.}, \bibinfo{year}{2020}.
\newblock \bibinfo{title}{smooth: Forecasting Using State Space Models}.
\newblock \URLprefix \url{https://CRAN.R-project.org/package=smooth}.
  \bibinfo{note}{{R} package version 2.6.0}.
%Type = Article
\bibitem[{Talagala et~al.(2019)Talagala, Li and Kang}]{talagala2019bmsr}
\bibinfo{author}{Talagala, T.}, \bibinfo{author}{Li, F.},
  \bibinfo{author}{Kang, Y.}, \bibinfo{year}{2019}.
\newblock \bibinfo{title}{{FFORMPP}: Feature-based forecast model performance
  prediction}.
\newblock \bibinfo{journal}{arXiv} \bibinfo{volume}{1908.11500}.
\newblock \URLprefix \url{https://arxiv.org/abs/1908.11500}.
%Type = Techreport
\bibitem[{Taylor and Letham(2017)}]{Taylor2017-lw}
\bibinfo{author}{Taylor, S.J.}, \bibinfo{author}{Letham, B.},
  \bibinfo{year}{2017}.
\newblock \bibinfo{title}{Forecasting at scale}.
\newblock \bibinfo{type}{Technical Report} \bibinfo{number}{e3190v2}. PeerJ
  Preprints.
%Type = Article
\bibitem[{Wang et~al.(2020)Wang, Zhou, Lu and Jiang}]{Wang2020-pc}
\bibinfo{author}{Wang, Z.}, \bibinfo{author}{Zhou, Z.}, \bibinfo{author}{Lu,
  H.}, \bibinfo{author}{Jiang, J.}, \bibinfo{year}{2020}.
\newblock \bibinfo{title}{Global and local sensitivity guided key salient
  object re-augmentation for video saliency detection}.
\newblock \bibinfo{journal}{Pattern Recognit.} \bibinfo{volume}{103},
  \bibinfo{pages}{107275}.
%Type = Article
\bibitem[{Wen et~al.(2017)Wen, Torkkola, Narayanaswamy and Madeka}]{Wen2017-ky}
\bibinfo{author}{Wen, R.}, \bibinfo{author}{Torkkola, K.},
  \bibinfo{author}{Narayanaswamy, B.}, \bibinfo{author}{Madeka, D.},
  \bibinfo{year}{2017}.
\newblock \bibinfo{title}{A {Multi-Horizon} quantile recurrent forecaster}
  \href{http://arxiv.org/abs/1711.11053}{{\tt arXiv:1711.11053}}.
%Type = Article
\bibitem[{Ye and Dai(2018)}]{Ye2018-ex}
\bibinfo{author}{Ye, R.}, \bibinfo{author}{Dai, Q.}, \bibinfo{year}{2018}.
\newblock \bibinfo{title}{A novel transfer learning framework for time series
  forecasting}.
\newblock \bibinfo{journal}{Knowl. Based Syst.} \bibinfo{volume}{156},
  \bibinfo{pages}{74--99}.
%Type = Incollection
\bibitem[{Yoon et~al.(2019)Yoon, Jarrett and van~der Schaar}]{Yoon2019-gr}
\bibinfo{author}{Yoon, J.}, \bibinfo{author}{Jarrett, D.},
  \bibinfo{author}{van~der Schaar, M.}, \bibinfo{year}{2019}.
\newblock \bibinfo{title}{Time-series generative adversarial networks}, in:
  \bibinfo{booktitle}{Advances in Neural Information Processing Systems 32}.
  \bibinfo{publisher}{Curran Associates, Inc.}, pp.
  \bibinfo{pages}{5508--5518}.
%Type = Article
\bibitem[{Yoon et~al.(2017)Yoon, Yun, Kim, Park and Jung}]{Yoon2017-tf}
\bibinfo{author}{Yoon, S.}, \bibinfo{author}{Yun, H.}, \bibinfo{author}{Kim,
  Y.}, \bibinfo{author}{Park, G.T.}, \bibinfo{author}{Jung, K.},
  \bibinfo{year}{2017}.
\newblock \bibinfo{title}{Efficient transfer learning schemes for personalized
  language modeling using recurrent neural network}
  \href{http://arxiv.org/abs/1701.03578}{{\tt arXiv:1701.03578}}.
%Type = Incollection
\bibitem[{Yosinski et~al.(2014)Yosinski, Clune, Bengio and
  Lipson}]{Yosinski2014-xz}
\bibinfo{author}{Yosinski, J.}, \bibinfo{author}{Clune, J.},
  \bibinfo{author}{Bengio, Y.}, \bibinfo{author}{Lipson, H.},
  \bibinfo{year}{2014}.
\newblock \bibinfo{title}{How transferable are features in deep neural
  networks?}, in: \bibinfo{editor}{Ghahramani, Z.}, \bibinfo{editor}{Welling,
  M.}, \bibinfo{editor}{Cortes, C.}, \bibinfo{editor}{Lawrence, N.D.},
  \bibinfo{editor}{Weinberger, K.Q.} (Eds.), \bibinfo{booktitle}{Advances in
  Neural Information Processing Systems 27}. \bibinfo{publisher}{Curran
  Associates, Inc.}, pp. \bibinfo{pages}{3320--3328}.
%Type = Article
\bibitem[{Yu et~al.(2017)Yu, Lin, Meng, Wei, Guo and Zhao}]{Yu2017-dm}
\bibinfo{author}{Yu, Y.}, \bibinfo{author}{Lin, H.}, \bibinfo{author}{Meng,
  J.}, \bibinfo{author}{Wei, X.}, \bibinfo{author}{Guo, H.},
  \bibinfo{author}{Zhao, Z.}, \bibinfo{year}{2017}.
\newblock \bibinfo{title}{Deep transfer learning for modality classification of
  medical images}.
\newblock \bibinfo{journal}{Information} \bibinfo{volume}{8},
  \bibinfo{pages}{91}.
%Type = Inproceedings
\bibitem[{Zhang et~al.(2018)Zhang, Kuppannagari, Kannan and
  Prasanna}]{Zhang2018-xn}
\bibinfo{author}{Zhang, C.}, \bibinfo{author}{Kuppannagari, S.R.},
  \bibinfo{author}{Kannan, R.}, \bibinfo{author}{Prasanna, V.K.},
  \bibinfo{year}{2018}.
\newblock \bibinfo{title}{Generative adversarial network for synthetic time
  series data generation in smart grids}, in: \bibinfo{booktitle}{2018 {IEEE}
  International Conference on Communications, Control, and Computing
  Technologies for Smart Grids ({SmartGridComm})}, pp. \bibinfo{pages}{1--6}.
%Type = Incollection
\bibitem[{Zhang et~al.(2015)Zhang, Zhao and LeCun}]{Zhang2015-vg}
\bibinfo{author}{Zhang, X.}, \bibinfo{author}{Zhao, J.},
  \bibinfo{author}{LeCun, Y.}, \bibinfo{year}{2015}.
\newblock \bibinfo{title}{Character-level convolutional networks for text
  classification}, in: \bibinfo{editor}{Cortes, C.}, \bibinfo{editor}{Lawrence,
  N.D.}, \bibinfo{editor}{Lee, D.D.}, \bibinfo{editor}{Sugiyama, M.},
  \bibinfo{editor}{Garnett, R.} (Eds.), \bibinfo{booktitle}{Advances in Neural
  Information Processing Systems 28}. \bibinfo{publisher}{Curran Associates,
  Inc.}, pp. \bibinfo{pages}{649--657}.
%Type = Article
\bibitem[{Zhuang et~al.(2018)Zhuang, Yan, Chen, Wang and Shen}]{Zhuang2018-nb}
\bibinfo{author}{Zhuang, N.}, \bibinfo{author}{Yan, Y.}, \bibinfo{author}{Chen,
  S.}, \bibinfo{author}{Wang, H.}, \bibinfo{author}{Shen, C.},
  \bibinfo{year}{2018}.
\newblock \bibinfo{title}{Multi-label learning based deep transfer neural
  network for facial attribute classification}.
\newblock \bibinfo{journal}{Pattern Recognit.} \bibinfo{volume}{80},
  \bibinfo{pages}{225--240}.
%Type = Incollection
\bibitem[{Zimmermann et~al.(2012)Zimmermann, Tietz and
  Grothmann}]{Zimmermann2012-cp}
\bibinfo{author}{Zimmermann, H.G.}, \bibinfo{author}{Tietz, C.},
  \bibinfo{author}{Grothmann, R.}, \bibinfo{year}{2012}.
\newblock \bibinfo{title}{Forecasting with recurrent neural networks: 12
  tricks}, in: \bibinfo{booktitle}{Neural Networks: Tricks of the Trade}.
  \bibinfo{publisher}{Springer, Berlin, Heidelberg}. Lecture Notes in Computer
  Science, pp. \bibinfo{pages}{687--707}.

\end{thebibliography}

\end{document}